\documentclass{article}
\usepackage[preprint]{corl_2022} 
\usepackage{graphicx} 
\usepackage{float} 
\usepackage{subfigure} 
\usepackage{wrapfig}
\usepackage{amssymb}
\usepackage{multirow}
\usepackage{rotating}
\usepackage{booktabs}
\usepackage{color}
\usepackage{amsmath}

\begin{document}

\title{MaskRange: A Mask-classification Model for Range-view based LiDAR Segmentation}
\author{Yi Gu, Yuming Huang, Chengzhong Xu, Hui Kong\\
       University of Macau\\
 \tt{\{yc17463, yc17907, huikong, czxu\}@um.edu.mo}
}


\maketitle


\begin{abstract}

Range-view based LiDAR segmentation methods are attractive for practical applications due to their direct inheritance from efficient 2D CNN architectures. In literature, most range-view based methods follow the per-pixel classification paradigm. Recently, in the image segmentation domain, another paradigm formulates segmentation as a mask-classification problem and has achieved remarkable performance. This raises an interesting question: can the mask-classification paradigm benefit the range-view based LiDAR segmentation and achieve better performance than the counterpart per-pixel paradigm? To answer this question, we propose a unified mask-classification model, MaskRange, for the range-view based LiDAR semantic and panoptic segmentation. Along with the new paradigm, we also propose a novel data augmentation method to deal with overfitting, context-reliance, and class-imbalance problems. Extensive experiments are conducted on the SemanticKITTI benchmark. Among all published range-view based methods, our MaskRange achieves state-of-the-art performance with $66.10$ mIoU on semantic segmentation and promising results with $53.10$ PQ on panoptic segmentation with high efficiency. Our code will be released.









\end{abstract}

\keywords{LiDAR segmentation, Mask-classification, Data augmentation, Autonomous vehicle and robot} 

\section{Introduction}\label{sec:introduction}
\begin{wrapfigure}[19]{R}{9cm}
	\centering
	\includegraphics[width=0.6\textwidth]{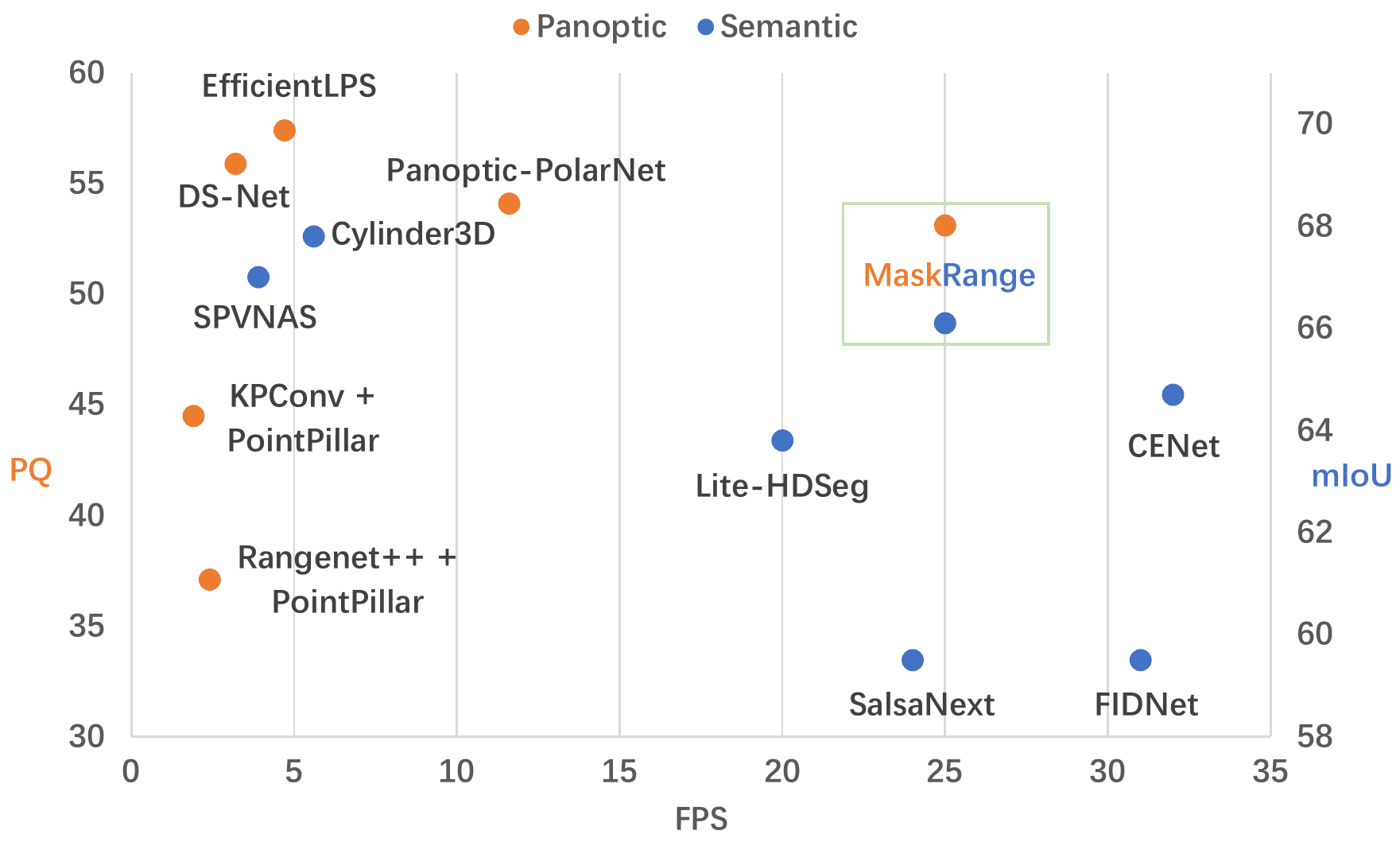}
	\caption{\footnotesize PQ and mIoU vs. single-frame inference latency on
		SemanticKITTI \cite{behley2019semantickitti}. Our MaskRange can achieve remarkable performance on both semantic and panoptic segmentation tasks with 25 FPS.}
	\label{intro_fig}
\end{wrapfigure}

Light Detection and Ranging (LiDAR) segmentation plays a significant role in autonomous driving and robot navigation by providing agents a full understanding of their surroundings. Following the development of image segmentation \cite{long2015fully}, most of the existing LiDAR-based methods \cite{guo2020deep, li2020deep, qi2017pointnet, milioto2019rangenet++} formulate LiDAR segmentation as a per-point classification problem. 

Very recently, in the image domain, some methods \cite{cheng2021per, cheng2021mask2former, zhang2021k} revive another paradigm, where image segmentation is formulated as a mask-classification problem by disentangling this task into two parallel branches: partitioning the image and classifying each partition. In a sense, the mask-classification paradigm is more anthropomorphic. Moreover, these methods have shown remarkable performance on both semantic- and instance-level segmentation tasks in a unified manner. 

This naturally raises an interesting question: can the mask-classification paradigm benefit LiDAR segmentation? To our best knowledge, no work has been investigated so far to answer this question. To fill this gap, we propose a novel mask-classification model for LiDAR-based segmentation, dubbed as MaskRange. We choose the range-view-based representation due to its efficiency and similarity to the image format. We build up our model based on the meta mask-classification architecture \cite{cheng2021per, cheng2021mask2former}, with the consideration of run-time and memory consumption. 

Intuitively, this adaptation to LiDAR's range-view representation should be readily effective for segmentation. However, due to the issues caused by overfitting, context reliance, and class imbalance, the performance is not as good as expected. The empirical results even cannot be on par with our per-pixel baseline. To tackle these problems, we propose a novel data augmentation scheme, named Weighted Paste Drop (WPD). The WPD takes negligible cost and can dramatically improve the performance of our model. In SemanticKITTI \cite{behley2019semantickitti} benchmark (accessed June 2022), our MaskRange achieves $66.10$ mIoU (rank 1 in range-view based methods) in semantic segmentation track and $53.10$ PQ in panoptic segmentation track. Notably, our model only takes $40$ $ms$ ($25$ FPS) for both tasks on a single NVIDIA RTX 2080Ti GPU. See Figure \ref{intro_fig} for better comparison.

It is worthwhile to highlight that, other than focusing on designing one particular network structure, our work aims at introducing a more general paradigm that has the potential of being applied to different LiDAR data representations. As a typical example, the range-view representation is adopted to demonstrate the superiority of the mask-classification paradigm. We hope this work can provide some insights into the mask-classification based LiDAR segmentation paradigm. In summary, our main contributions are stated as follows:

(i) To our best knowledge, MaskRange is the first LiDAR segmentation method based on the mask-classification paradigm. We will release our code and pre-trained models.

(ii) We propose a novel data augmentation method to reduce the adverse effects caused by the overfitting, context-reliance, and class-imbalance problems. 

(iii) Our method is a unified framework that can be applied to both semantic and panoptic segmentation tasks. Correspondingly, we propose a unified weighted focal loss to re-balance the training data. We achieve the best performance in range-view-based LiDAR semantic segmentation and promising results in panoptic segmentation in SemanticKITTI benchmark.

\section{Related Work}
\label{sec:citations}
\subsection{LiDAR Segmentation}
According to different LiDAR data representations, LiDAR-based segmentation algorithms can be categorized into three types \cite{xu2021rpvnet}, i.e., the point-, voxel-, and projection-based methods. The point-based methods \cite{qi2017pointnet, qi2017pointnet++, hu2019randla} can extract features directly from the raw point cloud, but have the inefficient neighbor-searching problem as well as computational and memory limitations. The voxel-based methods \cite{zhu2021cylindrical, Choy_2019, tang2020searching} partition the space into a finite number of regions to accelerate the feature extraction process, but suffer serious information loss when resolution is reduced. The projection-based methods \cite{milioto2019rangenet++, xu2020squeezesegv3, cortinhal2020salsanext, zhao2021fidnet, razani2021lite} also have information loss, but can benefit from some efficient 2D CNN architectures. Due to their efficiency, projection-based methods are more attractive for practical applications. Our work belongs to this category, specifically to the range-view-based methods. 

\subsection{Mask-classification Image Segmentation}
Different from the conventional per-pixel classification paradigm, the mask-classification methods formulate the segmentation problem from another perspective by predicting a set of binary masks and assigning a single class to each mask in parallel. Following DETR \cite{carion2020end} framework, Maskformer \cite{cheng2021per} uses a set of learnable tokens to query the dense embeddings for mask prediction and shows promising results on both semantic and panoptic segmentation tasks. Similarly, K-Net \cite{zhang2021k} also unifies segmentation tasks in the view of dot product between kernel and feature maps. Mask2Former \cite{cheng2021mask2former} further improves the components of MaskFormer and outperforms most specialized architectures on all considered tasks and datasets. Our work utilizes the MaskFormer \cite{cheng2021per} meta architecture and aims to achieve LiDAR segmentation in a unified manner. 

\subsection{3D Data Augmentation}
Data augmentation is an important way to enhance model performance by increasing the diversity of training sets. In the 3D area, simple data augmentation schemes such as random rotation, translation, point dropping, and flipping are widely used in common practice \cite{qi2017pointnet, qi2017pointnet++, aksoy2019salsanet}. PointMixup \cite{chen2020pointmixup} and PointCutMix \cite{zhang2021pointcutmix} extend the 2D augmentation methods Mixup \cite{DBLP:conf/iclr/ZhangCDL18} and CutMix \cite{DBLP:conf/iccv/YunHCOYC19} into 3D point cloud, respectively. These methods have shown better performance compared with the original simple augmentation methods. However, they cannot generally be applied to large-scale point cloud understanding. Mix3D \cite{nekrasov2021mix3d} proposes an effective method by simply mixing two data items and can significantly improve the out-of-context generalization ability. To alleviate the class-imbalance problem, RPVNet \cite{xu2021rpvnet} and Second \cite{DBLP:journals/sensors/YanML18} augment data by pasting the rare-class instances to the scene with the consideration of plausibility and reality. Our Weighted Paste Drop data-augmentation scheme is similar to these works but more efficient and unified. We will compare our method with the above-mentioned methods in subsection \ref{PasteMix}.


\section{MaskRange} \label{sec:methods}
The goal of LiDAR semantic segmentation is to predict the semantic label for each point. As for panoptic segmentation, points are classified into two types: things and stuff. A unique id is required for each point belonging to things \cite{li2021panoptic}. Unlike most methods proposed for a single task specially, in this study, we aim to achieve both tasks in a unified manner. 

We choose the range view as a proxy representation for LiDAR scans due to its efficiency and similarity to the image format. Generally, there are two ways to get the range-view representation: spherical projection \cite{milioto2019rangenet++} and scan unfolding \cite{triess2020scan}. Scan unfolding can recover a more dense representation but is not convenient to make data augmentation. Since a proper data augmentation is necessary to increase the diversity of training data, we choose the spherical projection to get the range image
\begin{equation}
\left(
\begin{array}{ccc}
    u\\
    v\\
    \end{array}
\right) = 
\left(
\begin{array}{ccc}
    \frac{1}{2}\left[1-arctan(y, x)\pi^{-1}\right]W\\
    \left[1-(arcsin(zr^{-1})+f_{up})\frac{1}{f}\right]H\\
\end{array}
\right),
\end{equation}
where W and H are the width and height of the range image, respectively. $f = f_{up} + f_{down}$ is the LiDAR’s vertical field-of-view. The range value $r=\sqrt{x^2+y^2+z^2}$ is calculated according to the point coordinates $\left[x, y, z\right]^T$ and $(u, v)^T$ are image coordinates in the range view. Following the previous works \cite{milioto2019rangenet++, zhao2021fidnet}, the input to our network is a $(H, W, 5)$ tensor $R_{in}$ with channels $(x, y, z, rem, r)$, where $rem$ is the remission
or intensity value.

\subsection{Architecture}
\begin{figure}[htbp]
	\centering
	\includegraphics[width=1.0\textwidth]{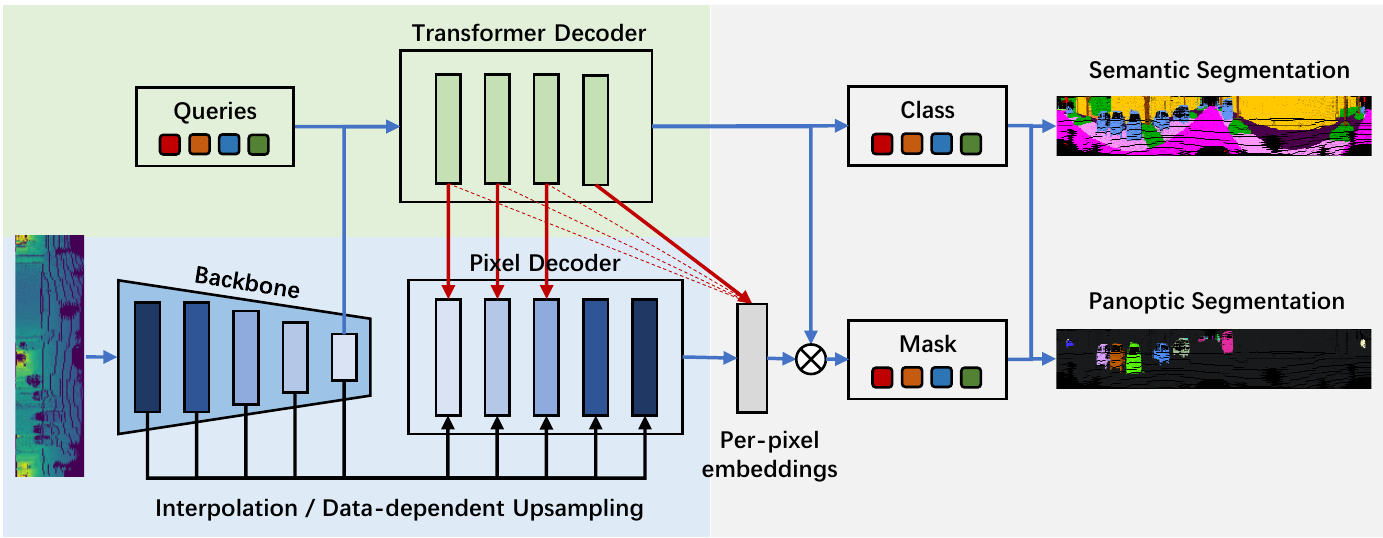} 
	\caption{The illustration of the MaskRange architecture. The dashed lines in red mean the auxiliary losses in original MaskFormer \cite{cheng2021per}, while the solid red lines are auxiliary losses by ours. This extra supervision makes our FID-like decoder perform better.} 
	\label{architecture}   
\end{figure}

We build up our model based on the mask-classification meta architecture \cite{cheng2021per, cheng2021mask2former} with consideration of the efficiency and accuracy. A simple meta architecture generally consists of three components: Backbone, Pixel Decoder and Transformer Decoder. Figure \ref{architecture} is the overview of MaskRange. Each component is introduced in detail in the following paragraphs.

\textbf{Backbone.} The backbone is to extract low-resolution features from the input $R_{in}$. Without loss of generality, we use a modified ResNet-34 \cite{he2016deep, cheng2022cenet, zhao2021fidnet} as our backbone to extract global features from a range image (see supplementary material Section A for more details). Note that we do not utilize any pre-trained parameters for two reasons. One is that we hope our method is general enough so that different types of backbones can be incorporated into it, whereas there may be no pre-trained models for these backbones. The other is that the ImageNet \cite{deng2009imagenet} pre-trained model is not suitable for our range image since the data distribution is quite different.

\textbf{Pixel Decoder.} The pixel decoder is used to get high-resolution per-pixel embeddings. Since we have no pre-trained models, our backbone is required to be trained from scratch and we also need to optimize the pixel decoder and transformer decoder simultaneously. Thus, it would take relatively long time to train the network. To speed up the feed-forward time and the convergence rate, we utilize the Fully Interpolation Decoding (FID) \cite{zhao2021fidnet} as our decoder module. The FID module is a parameter-free decoder, which only contains bilinear upsampling operation and can facilitate the optimization of our network. To further improve the performance, we replace the bilinear upsampling in FID with the Data-dependent Upsampling (DUpsampling) \cite{tian2019decoders} to get the high-resolution feature maps. Note that we design our decoder in FID format only for simplicity and efficiency. Other decoders can also be used.

\textbf{Transformer Decoder.} For DETR-like frameworks \cite{carion2020end, cheng2021per}, the transformer decoder acts like first asking what objects are in the image, then finding where these objects are. To achieve that, a set of learnable object embeddings are used to interact with the image features. The outputs of the transformer decoder are the corresponding class predictions and mask embeddings. The binary mask predictions can be obtained via a simple matrix multiplication operation. We do not incorporate more advanced designs (e.g., Mask2Former \cite{cheng2021mask2former}) into our architecture for simplicity. 
With this architecture, we can achieve the semantic segmentation and panoptic segmentation tasks in a unified manner. We refer readers to MaskFormer \cite{cheng2021per} for more details and better comparison. 

\subsection{Weighted Paste Drop Data Augmentation}  \label{PasteMix}
\vspace{5px}
To prevent model overfitting, the simple data augmentation schemes are widely used in common practice, including random rotation, translation, flipping, and point dropping \cite{cortinhal2020salsanext, zhao2021fidnet, razani2021lite}. We also train our model directly with these common data augmentation schemes. However, the performance of our MaskRange even cannot be on par with that of the per-pixel baseline (CENet \cite{cheng2022cenet}). We analyze that the unsatisfactory results (see supplementary material Section B for experimental analysis) are due to three problems: overfitting, context reliance, and class imbalance. 

\textbf{Overfitting.} We restrict the overfitting problem here only related to the amount and diversity of training data. It has been pointed out by some previous works \cite{vaswani2017attention, DBLP:conf/iclr/DosovitskiyB0WZ21} that transformer architectures usually require more training data compared with convolutional neural networks. However, the public available LiDAR segmentation datasets are relatively limited and small, and the common data augmentation schemes cannot satisfy the requirement of MaskRange.

\textbf{Context Reliance.} In terms of ``Context'', it generally refers to the strong configuration rules present in man-made environments \cite{nekrasov2021mix3d}. For example, with context knowledge, it can be inferred that traffic signs tend to be at the roadside, but not within a parking lot. However, relying too heavily on contextual cues may result in poor model generalization ability to rare or unseen situations \cite{nekrasov2021mix3d, DBLP:conf/cvpr/ShettySF19}. This problem is particularly acute for MaskRange since the transformer decoder only queries the global context embeddings to find objects. Intuitively, paying more attention to local geometrical information can be helpful to deal with the context-reliance problem. Unfortunately, most LiDAR segmentation methods adopt encoder-decoder architectures to aggregate context information without the consideration of balancing the global context priors and the local geometry information.   

\textbf{Class Imbalance.} Generally, the number of points corresponding to ``road" is significantly larger than that of ``pedestrian". Thus, the network is more biased to the high-frequency classes and may perform badly in some rare classes. This problem is widespread in LiDAR-based segmentation datasets \cite{milioto2019rangenet++} and can usually be dealt with by adding statistical weights to the cross-entropy loss, i.e., weighted cross entropy \cite{milioto2019rangenet++, cortinhal2020salsanext, zhao2021fidnet, razani2021lite}. However, this simple method cannot handle the class-imbalance problem well.

To deal with the aforementioned problems, we propose a novel data-augmentation method, which is constituted by three meta operations: ``Weighted'', ``Paste'' and ``Drop''. Initially, two frames are randomly selected from the dataset (noted as the first frame and the second frame) and the common data augmentation is applied. The \textbf{paste} operation selects the long-tail objects from the second frame firstly, then adds them to the first frame. The \textbf{drop} operation selects the non-long-tail class points in the first frame, then deletes these points. The \textbf{weighted} operation is to add a probability to the \textbf{paste} and \textbf{drop}.  

Our WPD data augmentation significantly enlarges the size and diversity of the dataset, therefore alleviating the overfitting problem. The core idea of our WPD scheme to alleviate the context-reliance problem is to weaken the role of the context priors. Our ``paste" operation can create unusual or even impossible scene scenarios to the training set. For example, we can ``paste" a “trunk" in the middle of a road, which never appears in the original dataset and cannot be created by common data augmentation. This means that the ``paste" operation can weaken the context priors. To further reduce the context bias, we ``drop" the points with high context information. For example, we hope our network can recognize cars without the road background. For the class-imbalance problem, we drop the high-frequency classes, such as road and car, with high probability and paste less-frequency classes frequently.

\begin{figure}[htb]
\centering
\subfigure[raw image, I.P. and our \textbf{paste}.]{\label{aug_1}
\includegraphics[width=0.32\textwidth]{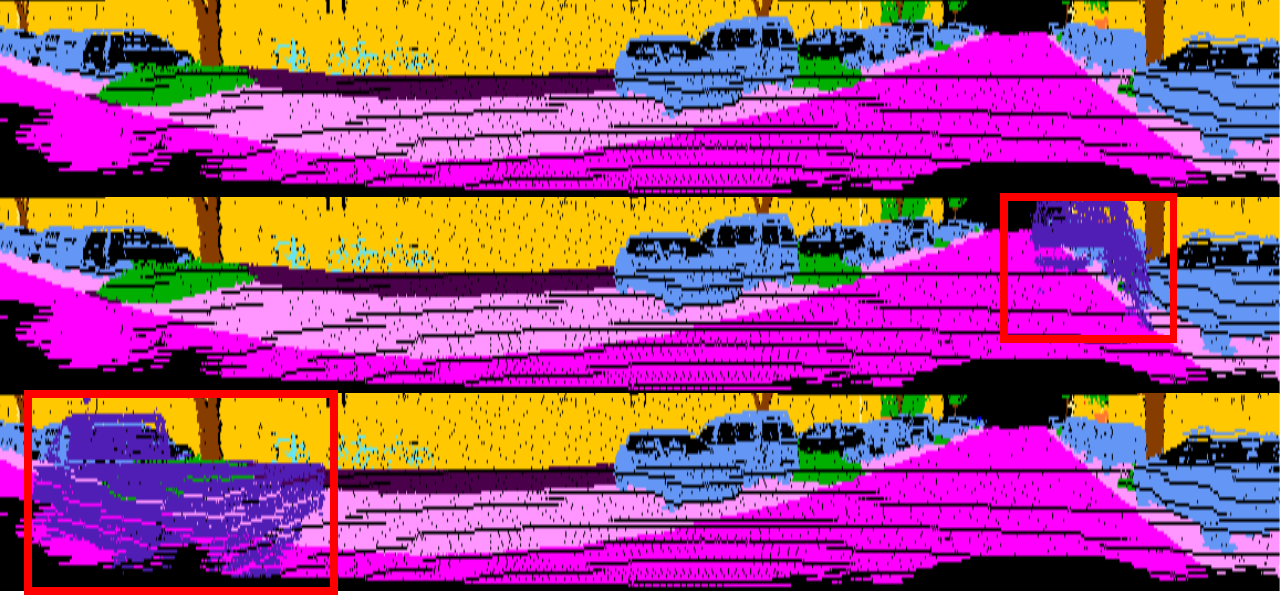}}
\subfigure[raw image, I.P. and our \textbf{paste}.]{\label{aug_2}
\includegraphics[width=0.32\textwidth]{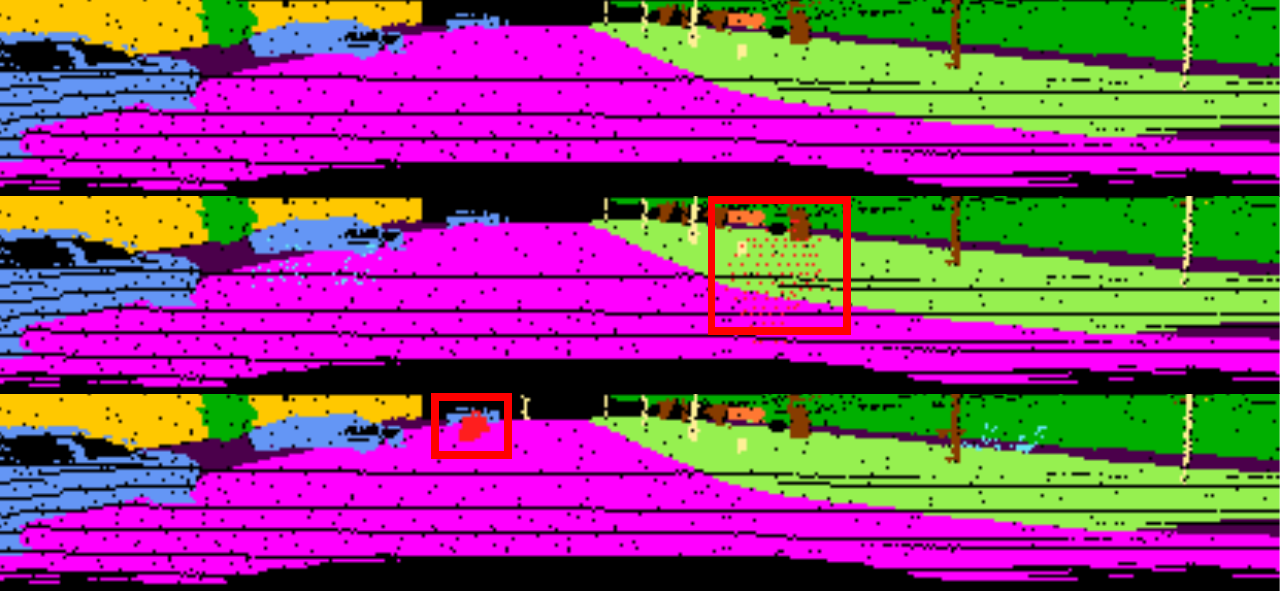}}
\subfigure[raw image, Mix3D, our \textbf{WPD}.]{\label{aug_3}
\includegraphics[width=0.32\textwidth]{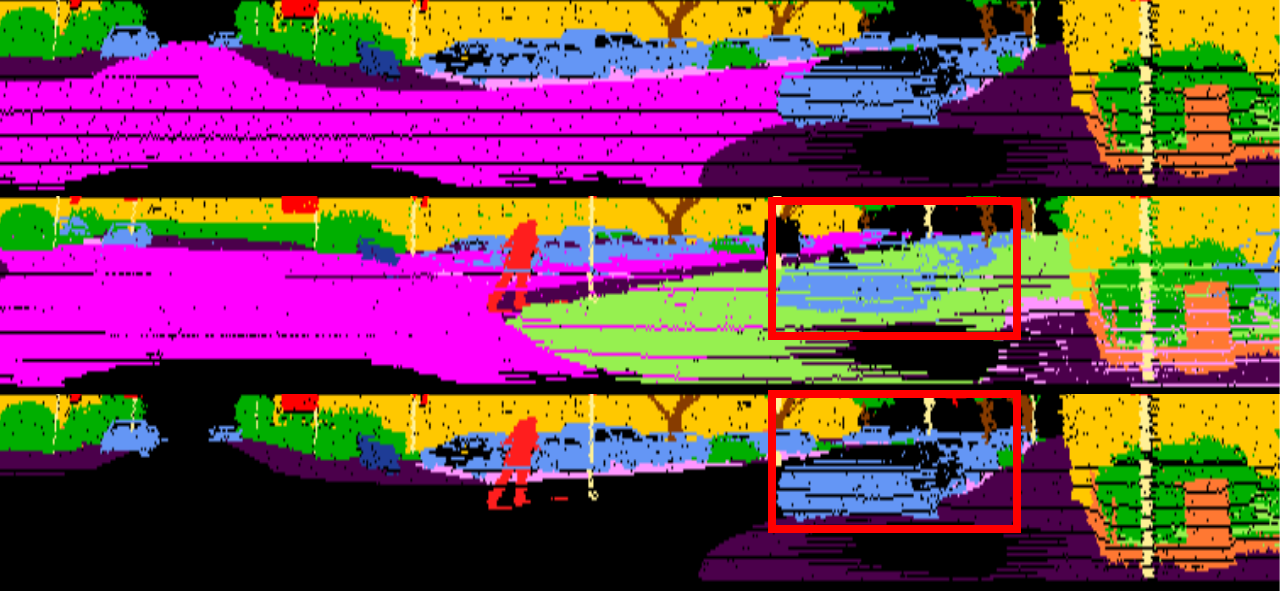}}
\caption{Qualitative comparisons of our \textbf{WPD} with Mix3D \cite{nekrasov2021mix3d} and Instance Paste \cite{xu2021rpvnet, DBLP:conf/cvpr/ShettySF19} (abbreviated as I.P.). Figure \ref{aug_1} and \ref{aug_2} show that our \textbf{paste} can better handle the orientation problem and the sparsity problem, respectively. Figure \ref{aug_3} shows that our \textbf{WPD} is more realistic compared with Mix3D. We highlight the augmented areas with red bounding box for better comparison.}
\label{mix}
\end{figure}
Note that, different from Mix3D \cite{nekrasov2021mix3d} which pastes all components, we only paste the long-tail objects and drop the non-long-tail objects. As Figure \ref{aug_3} shows, after the spherical projection, it is hard to distinguish the mixed points by Mix3D from the original points of the first frame. Though the context priors can be ``misled" by Mix3D \cite{nekrasov2021mix3d}, it makes the model confused. In addition, Mix3D cannot handle the class-imbalance problem. RPVNet \cite{xu2021rpvnet} and Second \cite{DBLP:journals/sensors/YanML18} also copy the long-tail instances from a database and paste them above the ground-class points. However, their methods are time-consuming and unrealistic (See Figure \ref{aug_1} and \ref{aug_2}). Our \textbf{paste} combines the actual scanning points, which inherently considers the orientation of the LiDAR sensor and the sparsity of the scanned points at different distances.

\subsection{Loss Function}
Our loss functions contain classification loss and binary mask losses. The mask-classification loss is the cross-entropy loss. The binary mask losses contain focal loss $L_{focal}$ \cite{lin2017focal}, Lov\'asz loss $L_{ls}$ \cite{berman2018lovasz} and boundary loss $L_{bd}$ \cite{bokhovkin2019boundary}. Our total loss can be computed as
\begin{equation}
    L = \lambda_1 L_{cls} + \lambda_2 L_{focal} + \lambda_3 L_{ls} + \lambda_4 L_{bd},
\end{equation}
where $\lambda_1$, $\lambda_2$, $\lambda_3$ and $\lambda_4$ are hyper parameters to balance these losses.

Different from other works \cite{cortinhal2020salsanext, zhao2021fidnet, razani2021lite, cheng2022cenet} which utilize the weighted cross-entropy loss, we utilize the weighted focal loss to re-balance the training set since the statistical weights have been largely changed by our data augmentation. Our weighted focal loss can be computed as 

\begin{equation}
    L_{focal}(p_t) = \alpha_i\beta_i(1-p_t)^\gamma\log(p_t),\ \  with\ \  \alpha_i = \frac{1}{f_i + \epsilon},\ \  and\ \  \beta_i =\frac{ins_i}{sem_i}, \label{loss_wfocal}
\end{equation}
where $p_t$ is the probability of prediction for the corresponding label. $\gamma$ is the focusing parameter and $(1-p_t)^\gamma$ is the modulating factor \cite{lin2017focal}. $f_i$ stands for the proportion of the class $i$ according to the number of points. $\epsilon$ is a small constant to prevent the exception value. $ins_i$ and $sem_i$ are the number of instance and semantic segments of the class $i$, respectively. Thus, $\alpha_i$ is the semantic re-balance factor and $\beta_i$ is the panoptic re-balance factor. For semantic segmentation, $\beta_i$ is equal to \textbf{1} since each class has only one ``instance". As for panoptic segmentation, $\beta_i$ is equal to \textbf{1} for stuff classes and larger than \textbf{1} for thing classes. This unified weighted focal loss provides a general promotion on both semantic and panoptic segmentation tasks.

In MaskFormer \cite{cheng2021per}, auxiliary losses are added to every transformer decoder layer and the per-pixel embeddings in the final pixel decoder layer. Our FID-like pixel decoder is not fully decoded, which may result in lower performance with only supervision on the final layer. Thus, inspired by deep supervision \cite{lee2015deeply, wang2015training, cheng2022cenet}, we add auxiliary losses on the feature embeddings in intermediate pixel decoder layers (as illustrated in Figure \ref{architecture}).


\section{Experiments}
\label{sec:experiments}
We evaluate our MaskRange on the public SemanticKITTI \cite{behley2019semantickitti} benchmark, which consists of 43551 LiDAR scans from 22 sequences. We use sequences 00 to 10 (except 08) as the training set and sequence 08 as the validation set, and present the implementation details in Sec. \ref{details}. We compare our results on the test set (sequences 11 to 21) with other state-of-the-art methods on both semantic (Sec. \ref{sem_seg}) and panoptic (Sec. \ref{pop_seg}) segmentation tasks. We conduct extensive ablation studies (Sec. \ref{ablation}) to analyze the effectiveness of each component of our method.

\subsection{Implementation Details} \label{details}
\textbf{Data augmentation.} Initially, two frames are randomly selected and the common data augmentation schemes (random point dropping, flipping, rotation, and translation) are applied to them. Then we \textbf{paste} the long-tail objects in the second frame to the first frame and \textbf{drop} the non-long-tail class points in the first frame. We set motorcyclist, bicyclist, bicycle, person, motorcycle, traffic-sign other-vehicle, truck, pole, other-ground, and trunk as long-tail classes according to the frequency of the points in each class (the detailed statistical results are presented in supplementary material Section C). The other classes are set as non-long-tail classes. We add the probability to the \textbf{paste} and \textbf{drop} operations, known as the \textbf{weighted} operation. We similarly obtain these weights as presented in our weighted focal loss Eq. \ref{loss_wfocal}, i.e., $\alpha_i\beta_i$. We rescale these weights to [0, 1] by $\alpha_i\beta_i$ / $\max$($\alpha_1\beta_1$, $\alpha_2\beta_2$, ..., $\alpha_k\beta_k$), where $k$ is the number of classes.

\textbf{Training settings.} The input resolution of our range image is $64\times2048$. CENet \cite{cheng2022cenet} is used as our pixel-level module, which provides us with the backbone and the pixel decoder. We use 4 transformer decoder layers with 100 queries by default. We use AdamW as the optimizer \cite{DBLP:conf/iclr/LoshchilovH19} and the poly \cite{ChenPKMY18} learning rate schedule to optimize our model. The initial learning rate of transformer decoder parameters is set as $0.0001$. For backbone and pixel decoder, the initial learning rate is $0.001$. The batch size is set as $16$. All models are trained for $150$ epochs with $4$ A100 GPUs. 

\textbf{Inference and post-processing.} In inference time, we use the same strategy as proposed in MaskFormer \cite{cheng2021per}. With the predictions in range view, we need to further project these predictions back to the 3D point cloud. Following some previous works \cite{milioto2019rangenet++, cortinhal2020salsanext, razani2021lite}, we employ a KNN post-processing to alleviate the ``shadow" effect. The inference latency is measured on a platform with an AMD Ryzen 9 3950X CPU and an RTX 2080Ti GPU.

\subsection{Semantic Segmentation}\label{sem_seg}

\vspace{-5px}
\begin{table*}[htp]
	\caption[CPNET]{Quantitative comparison on SemanticKITTI \cite{behley2019semantickitti} test set. We compare our MaskRange with some recent SOTA point/voxel-based and projection-based methods. The best results in the compared point/voxel- and projection-based methods are both highlighted by \textbf{bold} numbers separately. The second-best results in the compared projection-based methods are highlighted in \textcolor{blue}{blue} numbers. The FPS measurements with $\ast$ are tested by us with a single RTX 2080Ti GPU.}
	\Huge
	\resizebox{0.98\columnwidth}{!}{
		\begin{tabular}
			{c|l|c|ccccccccccccccccccc|c}
			\toprule
			Category & Method
			& \begin{sideways} Mean IoU \end{sideways} 
			& \begin{sideways} Car \end{sideways} 
			& \begin{sideways} Bicycle \end{sideways} 
			& \begin{sideways} Motorcycle \end{sideways} 
			& \begin{sideways} Truck \end{sideways} 
			& \begin{sideways} Other-vehicle \end{sideways} 
			& \begin{sideways} Person \end{sideways} 
			& \begin{sideways} Bicyclist \end{sideways} 
			& \begin{sideways} Motorcyclist \end{sideways} 
			& \begin{sideways} Road \end{sideways} 
			& \begin{sideways} Parking \end{sideways} 
			& \begin{sideways} Sidewalk \end{sideways} 
			& \begin{sideways} Other-ground \end{sideways} 
			& \begin{sideways} Building \end{sideways} 
			& \begin{sideways} Fence \end{sideways} 
			& \begin{sideways} Vegetation \end{sideways} 
			& \begin{sideways} Trunk \end{sideways} 
			& \begin{sideways} Terrain \end{sideways} 
			& \begin{sideways} Pole \end{sideways} 
			& \begin{sideways} Traffic-sign \end{sideways} 
			& \begin{sideways} FPS (Hz) \end{sideways} \\
			\midrule
			\multirow{6}{*}{\begin{sideways} Point / voxel  \end{sideways} }
			& RandLA-Net \cite{hu2019randla} & 53.9 & 94.2 & 26.0 & 25.8 & 40.1 & 38.9 & 49.2 & 48.2 & 7.2 & 90.7 & 60.3 & 73.7 & 20.4 & 86.9 & 56.3 & 81.4 & 61.3 & 66.8 & 49.2 & 47.7 & 1.9  \\
			& KPConv \cite{thomas2019kpconv} & 58.8 & 96.0 & 30.2 & 42.5 & 33.4 & 44.3 & 61.5 & 61.6 & 11.8 & 88.8 & 61.3 & 72.7 & 31.6 & 90.5 & 64.2 & 84.8 & 69.2 & 69.1 & 56.4 & 47.4 & - \\
			& KPRNet   \cite{kochanov2020kprnet}
			& 63.1 & 95.5 & 54.1 & 47.9 & 23.6 & 42.6 & 65.9 & 65.0 & 16.5 & \textbf{93.2} & \textbf{73.9} & \textbf{80.6} & 30.2 & 91.7 & 68.4 & 85.7 & 69.8 & \textbf{71.2} & 58.7 & 64.1 & 0.3 \\
			& JS3C-Net \cite{yan2021sparse}
			& 66.0 & 95.8 & 59.3 & 52.9 & 54.3 & 46.0 & 69.5 & 65.4 & 39.9 & 88.9 & 61.9 & 72.1 & 31.9 & \textbf{92.5} & \textbf{70.8} & 84.5 & 69.8 & 67.9 & 60.7 & \textbf{68.7} & 2.1 (V100)\\
			& SPVNAS \cite{tang2020searching} & 67.0 & \textbf{97.2} & 50.6 & 50.4 & 56.6 & 58.0 & 67.4 & 67.1 & \textbf{50.3} & 90.2 & 67.6 & 75.4 & 21.8 & 91.6 & 66.9 & \textbf{86.1} & \textbf{73.4} & 71.0 & \textbf{64.3} & 67.3 & 3.9 (1080Ti) \\
			& Cylinder3D \cite{zhu2021cylindrical} & \textbf{67.8} & 97.1 & \textbf{67.6} & \textbf{64.0} & \textbf{59.0} & \textbf{58.6} & \textbf{73.9} & \textbf{67.9} & 36.0 & 91.4 & 65.1 & 75.5 & \textbf{32.3} & 91.0 & 66.5 & 85.4 & 71.8 & 68.5 & 62.6 & 65.6 & 5.6  \\
			\midrule
			\multirow{7}{*}{\begin{sideways} Projection  \end{sideways} }
			& RangeNet++ \cite{milioto2019rangenet++} & 52.2 & 91.4 & 25.7 & 34.4 & 25.7 & 23.0 & 38.3 & 38.8 & 4.8 & \textcolor{blue}{91.8} & 65.0 & 75.2 & 27.8 & 87.4 & 58.6 & 80.5 & 55.1 & 64.6 & 47.9 & 55.9 & 12 \\
			& SqueezeSegv3 \cite{xu2020squeezesegv3} & 55.9 & 92.5 & 38.7 & 36.5 & 29.6 & 33.0 & 45.6 & 46.2 & 20.1 & 91.7 & 63.4 & 74.8 & 26.4 & 89.0 & 59.4 & 82.0 & 58.7 & 65.4 & 49.6 & 58.9 & 6  \\
			& SalsaNext \cite{cortinhal2020salsanext} & 59.5 & 91.9 & 48.3 & 38.6 & {38.9} & 31.9 & 60.2 & 59.0 & 19.4 & 91.7 & 63.7 & 75.8 & 29.1 & 90.2 & 64.2 & 81.8 & 63.6 & 66.5 & 54.3 & 62.1 & 24 \\
			& FIDNet \cite{zhao2021fidnet}
			& 59.5 & \textcolor{blue}{93.9} & {54.7} & 48.9 & 27.6 & 23.9 & {62.3} & 59.8 & 23.7 & 90.6 & 59.1 & 75.8 & 26.7 & 88.9 & 60.5 & \textcolor{blue}{84.5} & 64.4 & {69.0} & 53.3 & 62.8 &  31* \\
			& Lite-HDSeg \cite{razani2021lite} 
			& {63.8} & 92.3 & 40.0 & \textcolor{blue}{55.4} & 37.7 & {39.6} & 59.2 & \textbf{71.6} & \textbf{54.1} & \textbf{93.0} & \textcolor{blue}{68.2} & \textbf{78.3} & {29.3} & \textbf{91.5} & {65.0} & 78.2 & {65.8} & 65.1 & {59.5} & \textbf{67.7} & 20  \\
			& CENet \cite{cheng2022cenet}
			& \textcolor{blue}{64.7} & 91.9 & \textbf{58.6} & 50.3 & \textcolor{blue}{40.6} & \textcolor{blue}{42.3} & \textbf{68.9} & \textcolor{blue}{65.9} & \textcolor{blue}{43.5} & 90.3 & 60.9 & 75.1 & \textbf{31.5} & \textcolor{blue}{91.0} &  \textbf{66.2} &  \textcolor{blue}{84.5} &  \textbf{69.7} & \textbf{70.0} & \textbf{61.5} & \textcolor{blue}{67.6} & 32*  \\
			\midrule
			& \textbf{MaskRange (ours)}
			& \textbf{66.1} & \textbf{94.2} & \textcolor{blue}{56.0} & \textbf{55.7} & \textbf{59.2} & \textbf{52.4} & \textcolor{blue}{67.6} & {64.8} & {31.8} & 91.7 & \textbf{70.7} & \textcolor{blue}{77.1} & \textcolor{blue}{29.5} & {90.6} & \textcolor{blue}{65.2} & \textbf{84.6} & \textcolor{blue}{68.5} & \textcolor{blue}{69.2} & \textcolor{blue}{60.2} & {66.6} & 25* \\
			\bottomrule
		\end{tabular}
	}
	\label{semantic_testset}
\end{table*}
\vspace{-5px}

For semantic segmentation, we use mean Intersection over Union (mIoU) \cite{everingham2015pascal} as our evaluation metric to compare our MaskRange with recently published works. As shown in Table \ref{semantic_testset}, among all projection-based methods, our MaskRange achieves the best performance. Compared with point/voxel-based methods, our performance is still competitive with only $40\ ms$ runtime cost.

\subsection{Panoptic Segmentation}\label{pop_seg}

\vspace{-5px}
\begin{table*}[htp]
\caption{Comparison of LiDAR panoptic segmentation on SemanticKITTI test set.}
\Huge
\footnotesize
\resizebox{0.98\columnwidth}{!}{
\begin{tabular}
{l|cccc|ccc|ccc|c|c}
\toprule
Method & PQ & PQ$^\dagger$ & RQ & SQ & PQ\textsuperscript{Th} & RQ\textsuperscript{Th} & SQ\textsuperscript{Th} & PQ\textsuperscript{St} & RQ\textsuperscript{St} & SQ\textsuperscript{St} & mIoU & FPS (Hz) \\
\midrule
RangeNet++ \cite{milioto2019rangenet++} + PP \cite{lang2019pointpillars} & 37.1 & 45.9 & 47.0 & 75.9 & 20.2 & 25.2 & 75.2 & 49.3 & 62.8 & 76.5 & 52.4 & 2.4 \\
LPSAD \cite{9340837} & 38.0 & 47.0 & 48.2 & 76.5 & 25.6 & 31.8 & 76.8 & 47.1 & 60.1 & 76.2 & 50.9 & 11.8 \\
KPConv \cite{thomas2019kpconv} + PP \cite{lang2019pointpillars} & 44.5 & 52.5 & 54.4 & 80.0 & 32.7 & 38.7 & 81.5 & 53.1 & 65.9 & 79.0 & 58.8 & 1.9 \\
Panoster \cite{gasperini2020panoster} & 52.7 & 59.9 & 64.1 & 80.7 & 49.4 & 58.5 & 83.3 & 55.1 & 68.2 & 78.8 & 59.9 & - \\
Panoptic-PolarNet \cite{zhou2021panoptic} & 54.1 & 60.7 & 65.0 & 81.4 & 53.3 & 60.6 & 87.2 & 54.8 & 68.1 & 77.2 & 59.5 & 11.6 \\
DS-Net \cite{hong2021lidar} & 55.9 & 62.5 & 66.7 & 82.3 & \textbf{55.1} & \textbf{62.8} & 87.2 & 56.5 & 69.5 & 78.7 & 61.6 & 3.2 \\
EfficientLPS \cite{sirohi2021efficientlps} & \textbf{57.4} & \textbf{63.2} & \textbf{68.7} & \textbf{83.0} & 53.1 & 60.5 & \textbf{87.8} & \textbf{60.5} & \textbf{74.6} & \textbf{79.5} & 61.4 & 4.7 (Titan RTX) \\
\midrule
\textbf{MaskRange (ours)} & 53.1 & 59.2  & 64.6 & 81.2 & 44.9 & 53.0 & 83.5 & 59.1 & 73.1 & \textbf{79.5} & \textbf{61.8} & 25\\
\bottomrule
\end{tabular}
}
\label{tab:panoptic_testset}
\end{table*}
\vspace{-5px}

For panoptic segmentation, we use Panoptic Quality (PQ) \cite{kirillov2019panoptic} as the main metric and we also report Recognition Quality (RQ) and Segmentation Quality (SQ) for better comparison. These metrics are calculated separately for thing and stuff classes. 
Besides, we also report PQ$^\dagger$ \cite{porzi2019seamless} which uses IoU as PQ for stuff classes. 
As shown in Table \ref{tab:panoptic_testset}, among all listed methods, our MaskRange is the most efficient one with 25 FPS and achieves comparable performance with $53.1$ PQ. Benefiting from the mask classification paradigm and our unified WPD augmentation, we can extend our MaskRange to panoptic segmentation without any bounding-box supervision or extra hyperparameters tuning.

\subsection{Ablation Study}\label{ablation}
We do ablation studies on the WPD data augmentation, upsampling and deep supervision strategies. We also compare our weighted focal loss (WF) with the weighted cross-entropy loss (WCE) to show the effectiveness of our loss functions. We choose CENet \cite{cheng2022cenet} as the per-pixel baseline and train it with the official code. All the ablation studies are conducted with the $64\times2048$ resolution.

\textbf{Data augmentation.} 
We first study the effectiveness of our WPD data augmentation. To show the superiority of our WPD, we also compare our method with Mix3D \cite{nekrasov2021mix3d} and Instance Paste (I.P.) \cite{xu2021rpvnet}. As shown in Table \ref{ablation_tale} (row 3-6), our WPD outperforms compared methods by at least $1.2$ mIoU. It can be seen that, with the common data augmentation, our model cannot even be on par with the CENet \cite{cheng2022cenet} (row 1). When incorporating the Mix3D \cite{nekrasov2021mix3d}, the performance is improved by a large margin. This motivates us to investigate better data augmentation methods. Thus, we propose WPD, which can alleviate the overfitting, context-reliance, and class-imbalance problems simultaneously. We refer readers to the supplementary materials Section B and D for more details. 

We also add our WPD to the CENet (row 2), but the improvement is limited ($64.25\%-62.70\%=1.55\%$). In contrast, the improvement attributed to the WPD in our MaskRange is $66.80\%-59.10\%=7.70\%$. This means that our MaskRange is more expressive. Compared with CENet, MaskRange introduces $1.113$ M extra parameters and only $2.2$ more GFlops. Another possible
explanation is that the transformer decoder collects global information from the image features to generate class predictions. This setup reduces the need of heavy context aggregation in per-pixel module \cite{cheng2021per} and can help this module focus more on the shape or geometry information.

\textbf{WCE vs. WF.} Since the statistical weights have been largely changed by our WPD data augmentation, we use focal loss to re-weight the cross-entropy loss. Additionally, we also introduce extra re-balance factors to the original focal loss. Rows $6$-$7$ of Table \ref{ablation_tale} show that our weighted focal loss is superior to the weighted cross-entropy loss.

\textbf{Deep supervision strategy.}
We compare our improved implementation of deep supervision (notated as DeepB) with the default deep supervision (DeepA). As illustrated in Figure \ref{architecture}, the red dashed lines represent DeepA and the red solid lines represent DeepB. As shown in row $7$ and $8$ of Table \ref{ablation_tale}, the mIoU of DeepB can be slightly higher than DeepA. Moreover, the convergence rate of DeepB is faster than that of DeepA. Thus, We choose DeepB as our deep supervision strategy.

\textbf{Upsampling strategy.}
The interpolation-based upsamling (IU) is efficient but may result in coarse predictions. With data-dependent upsampling (DU), we can achieve better performance by $0.24$ mIoU (row $8$-$9$) with only $2\ ms$ extra runtime cost.


We also do the ablation on the re-balance strategies, which contain the comparisons of no balance, class balance, and unified balance strategies on both semantic and panoptic segmentation tasks. We refer readers to the supplementary material Section E for more details.

\vspace{-10px}
\begin{table*}[htp]
\caption{Ablation studies of LiDAR semantic segmentation on SemanticKITTI validation set.}
\footnotesize
\resizebox{0.98\columnwidth}{!}{
\begin{tabular}{c|c c c c|c c|c c|c c|c}
\toprule
 & common & Mix & I.P. & WPD & WCE & WF & DeepA & DeepB & IU & DU & mIoU (\%)\\
\midrule
\multirow{2}*{CENet \cite{cheng2022cenet}} 
& \checkmark & & & &\checkmark & &  & & \checkmark& & 62.70\\
& & & & \checkmark &\checkmark & &  & & \checkmark& & 64.25\\
\midrule
\multirow{7}*{MaskRange} 
& \checkmark & & & &\checkmark & & \checkmark & & \checkmark& & 59.10\\
& & \checkmark & & &\checkmark & & \checkmark & & \checkmark& & 65.60\\
& & & \checkmark & &\checkmark & & \checkmark & & \checkmark& & 63.32\\
& & & & \checkmark & \checkmark & & \checkmark & & \checkmark& & 66.80\\
& & & & \checkmark & & \checkmark & \checkmark & & \checkmark& & 67.41\\
& & & & \checkmark & & \checkmark & & \checkmark& \checkmark& & 67.53\\
& & & & \checkmark & & \checkmark & & \checkmark& & \checkmark& 67.77\\
\bottomrule
\end{tabular}
}
\label{ablation_tale}
\end{table*}


\section{Conclusion}\label{sec:conclusion}
In this work, we investigate a unified LiDAR segmentation method (MaskRange) based on the mask-classification paradigm with the range-view LiDAR representation. With a novel Weighted Paste Drop data augmentation, our MaskRange achieves state-of-the-art performance on semantic segmentation and promising results on panoptic segmentation with high efficiency. Our method is flexible and general enough that can be readily applicable to other LiDAR segmentation methods. 







\bibliography{example}  

\clearpage
\appendix
\part*{Supplementary Materials}

\section{Network Details}
\begin{table}[htb]
  \centering
  \resizebox{0.8\columnwidth}{!}{
  \begin{tabular}[t]{l}
\begin{tabular}[t]{|l|l|l|l|l|l|l|l|}
\hline
\multicolumn{8}{|l|}{\textbf{Backbone}} \\
\hline
\textbf{layer} & \textbf{k} & \textbf{s} & \textbf{chns\_in} & \textbf{chns\_out} & \textbf{output} & \textbf{output size}   & \textbf{activation}    \\ \hline
conv1  & 3      & 1      & 5      & 64  &  -  & $64\times2048$ & LeakyReLU \\
conv2  & 3      & 1      & 64     & 128 &  -   & $64\times2048$ & LeakyReLU \\
conv3  & 3      & 1      & 128    & 128 &  $x_0$   & $64\times2048$ & LeakyReLU \\ \hline
Basic block $\times3$ & 3  & 1 & 128 & 128 &  $x_1$ & $64\times2048$ & LeakyReLU \\ \hline
Basic block $\times4$ & 3  & 2 & 128 & 128 &  $x_2$ & $32\times1024$ & LeakyReLU \\ \hline
Basic block $\times6$ & 3  & 2 & 128 & 128 &  $x_3$ & $16\times512$ & LeakyReLU \\ \hline
Basic block $\times3$ & 3  & 2 & 128 & 128 &  $x_4$ & $\ \ 8\times256$  & LeakyReLU \\
\hline
\end{tabular} \\
\begin{tabular}[t]{|l|l|l|l|l|l|l|}

\hline
\multicolumn{5}{|l|}{\textbf{Pixel Decoder}} \\
\hline
\textbf{operation} & \textbf{chns} & \textbf{input} & \textbf{output} & \textbf{resolution}\\ \hline
$\uparrow$  & 128   & $x_2$ & $x_2\uparrow$   & $64\times2048$ \\
$\uparrow$  & 128   & $x_3$ & $x_3\uparrow$   & $64\times2048$ \\
$\uparrow$  & 128   & $x_4$ & $x_4\uparrow$   & $64\times2048$ \\
\hline
Concatenate & 640   & $(x_0, x_1, x_2\uparrow, x_3\uparrow, x_4\uparrow)$ & cat   & $64\times2048$ \\ \hline 
Conv2d      & 256   & cat   & embeddings      & $64\times2048$ \\ \hline
Conv2d      & 128   & embeddings & pixel\_embeddings & $64\times2048$ \\ \hline

\end{tabular} \\

\begin{tabular}[t]{|l|l|l|l|l|}

\hline
\multicolumn{5}{|l|}{\textbf{Transformer Decoder}} \\
\hline
\textbf{layer} & \textbf{operations} &\textbf{input} & \textbf{output} & \textbf{activation} \\ \hline
layer1 & self-attention, cross-attention, FFN & $x_4$, querys  & $o_1$  & ReLU \\ \hline
layer2 & self-attention, cross-attention, FFN & $o_1$, querys  & $o_2$  & ReLU \\ \hline 
layer3 & self-attention, cross-attention, FFN & $o_2$, querys  & $o_3$  & ReLU \\ \hline
layer4 & self-attention, cross-attention, FFN & $o_3$, querys  & $o_4$  & ReLU \\ \hline
\end{tabular} 

\end{tabular} 
 }
  \caption{\textbf{MaskRange.} Here \textbf{k} is the kernel size. \textbf{s} is the stride. \textbf{chns\_in} and \textbf{chns\_out} are the number of input and output channels. $\uparrow$ is the interpolation or data-dependent upsampling operation.}
\label{tab:network}
\end{table}

\section{Overfitting, Class Imbalance and Context Reliance Analysis}
\textbf{Overfitting.} We train our MaskRange and CENet \cite{cheng2022cenet} with the same common data augmentation. Table \ref{over_appendix} shows that the performance of MaskRange in trainset is better than that of CENet. However, it is worse in validation set. This is a typical overfitting phenomenon. With our WPD, the performance gap is reduced.

\begin{table}[h!]
\caption{Best performance on the train and validation set}
\centering
 \begin{tabular}{|c|c|c|} 
 \hline
 Methods & train set & validation set \\ [0.5ex] 
 \hline
 CENet \cite{cheng2022cenet} & 66.8 & 62.7  \\ 
 \hline
 MaskRange & 89.6 & 59.1  \\
 \hline
 MaskRange + WPD & 84.6 & 67.7  \\
 \hline
 \end{tabular}
\label{over_appendix}
\end{table}

\textbf{Class Imbalance.}  

\begin{table*}[htp]
\caption[CPNET]{Class-wise comparison on SemanticKITTI \cite{behley2019semantickitti} validation set.}
\Huge
\resizebox{0.98\columnwidth}{!}{
\begin{tabular}
{|l|c|ccccccccccccccccccc|}
\toprule
Method
& \begin{sideways} Mean IoU \end{sideways} 
& \begin{sideways} Car \end{sideways} 
& \begin{sideways} Bicycle \end{sideways} 
& \begin{sideways} Motorcycle \end{sideways} 
& \begin{sideways} Truck \end{sideways} 
& \begin{sideways} Other-vehicle \end{sideways} 
& \begin{sideways} Person \end{sideways} 
& \begin{sideways} Bicyclist \end{sideways} 
& \begin{sideways} Motorcyclist \end{sideways} 
& \begin{sideways} Road \end{sideways} 
& \begin{sideways} Parking \end{sideways} 
& \begin{sideways} Sidewalk \end{sideways} 
& \begin{sideways} Other-ground \end{sideways} 
& \begin{sideways} Building \end{sideways} 
& \begin{sideways} Fence \end{sideways} 
& \begin{sideways} Vegetation \end{sideways} 
& \begin{sideways} Trunk \end{sideways} 
& \begin{sideways} Terrain \end{sideways} 
& \begin{sideways} Pole \end{sideways} 
& \begin{sideways} Traffic-sign \end{sideways} \\
\midrule
MaskRange + common\_aug
& {59.10} & 95.94 & 14.53 & 39.00 & 60.89 & 48.96 & 68.00 & 73.84 & 0.0 & 95.09 & 47.13 & 82.95 & 0.15 & 89.50 & 60.79 & 87.46 & 69.12 & 75.35 & 64.65 & 48.85\\
\midrule
MaskRange + WPD
& {67.77} & {97.21} & {59.08} & {67.46} & {87.33} & {65.87} & {79.37} & {89.82} & {4.19} & 95.65 & {49.68} & {84.01} & {12.08} & {89.85} & {61.78} & {85.94} & {70.40} & {72.06} & {66.03} & {49.81}\\
\bottomrule
\end{tabular}
}
\label{appendix_imbalance}
\end{table*}

We analyze the imbalance problem from the empirical perspective in this section. We compare the class-wise performance of MaskRange with common data augmentation and WPD and present the results in Table \ref{appendix_imbalance}. Since the other ground and motorcyclist classes are seldom present in the training set, no matter what kind of augmentation methods we use, MaskRange still performs badly in these two classes. This extreme example illustrates the imbalance problem. Table \ref{appendix_imbalance} also shows the effectiveness of our method, where the performances of the bicycle and motorcycle classes are improved at least $28$ mIoU.

\begin{figure}[htbp]
    \centering
    \subfigure[]{\label{sup_1}
    \includegraphics[width=0.32\textwidth]{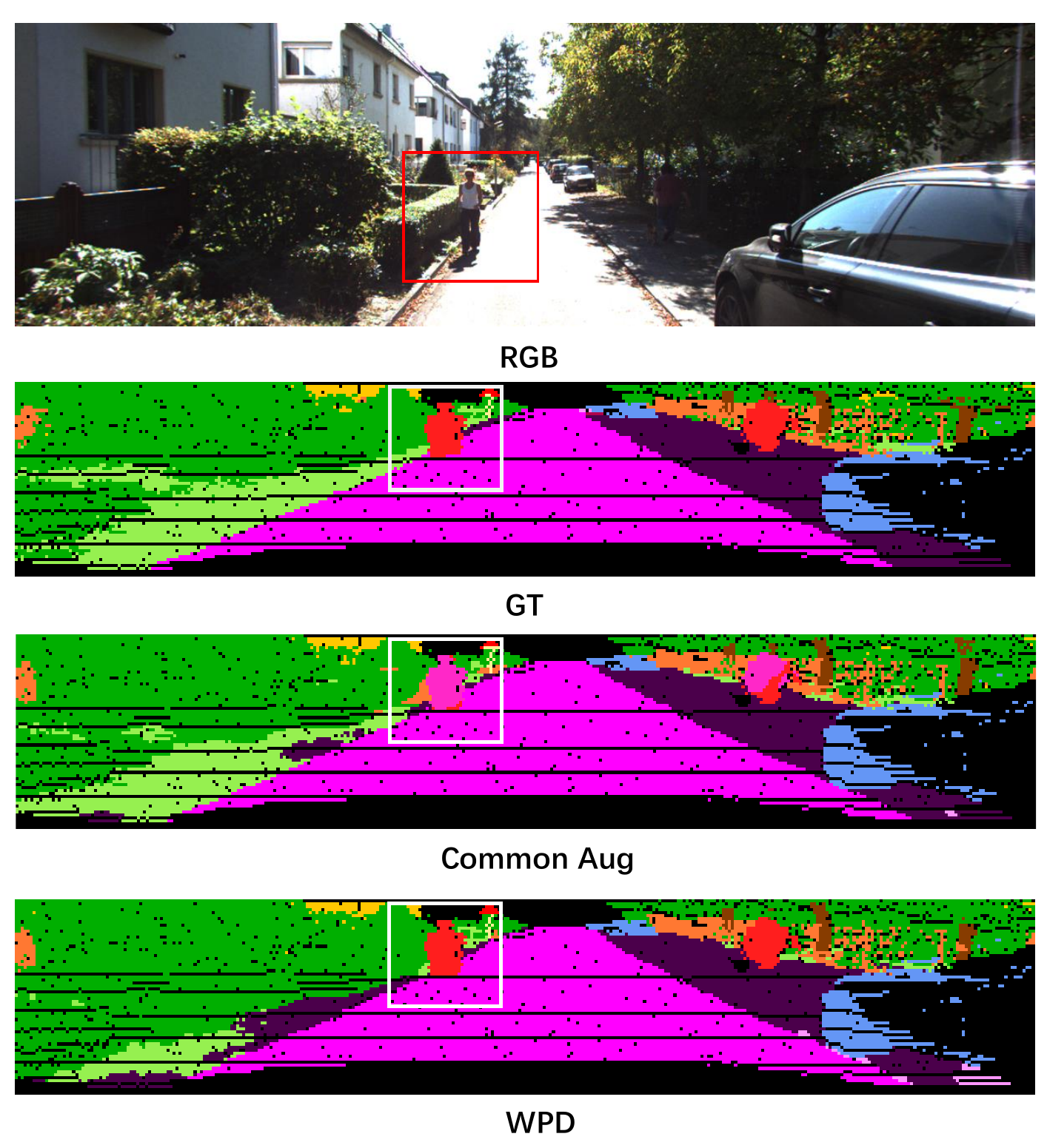}}
    \subfigure[]{\label{sup_2}
    \includegraphics[width=0.32\textwidth]{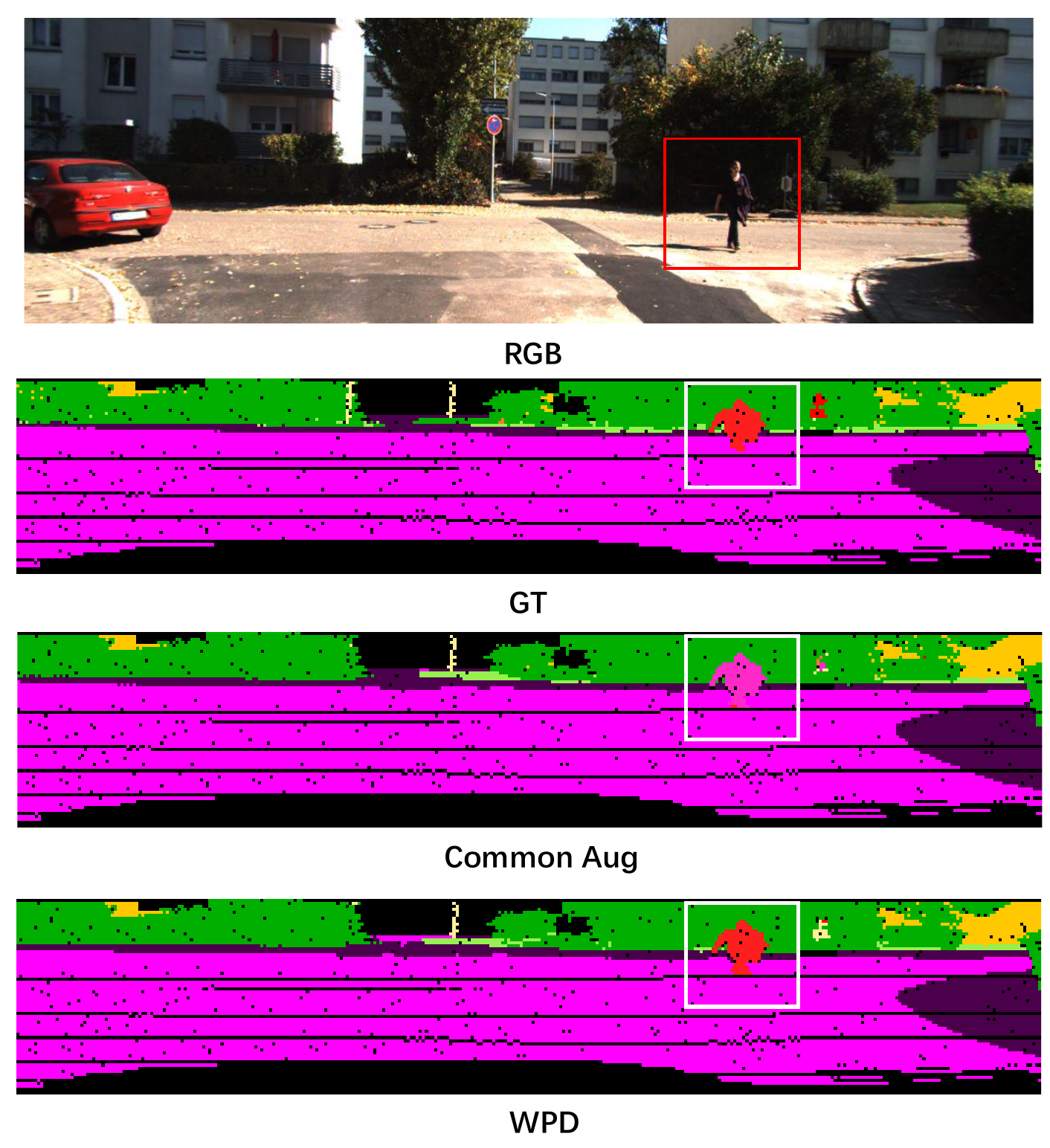}}
    \subfigure[]{\label{sup_3}
    \includegraphics[width=0.32\textwidth]{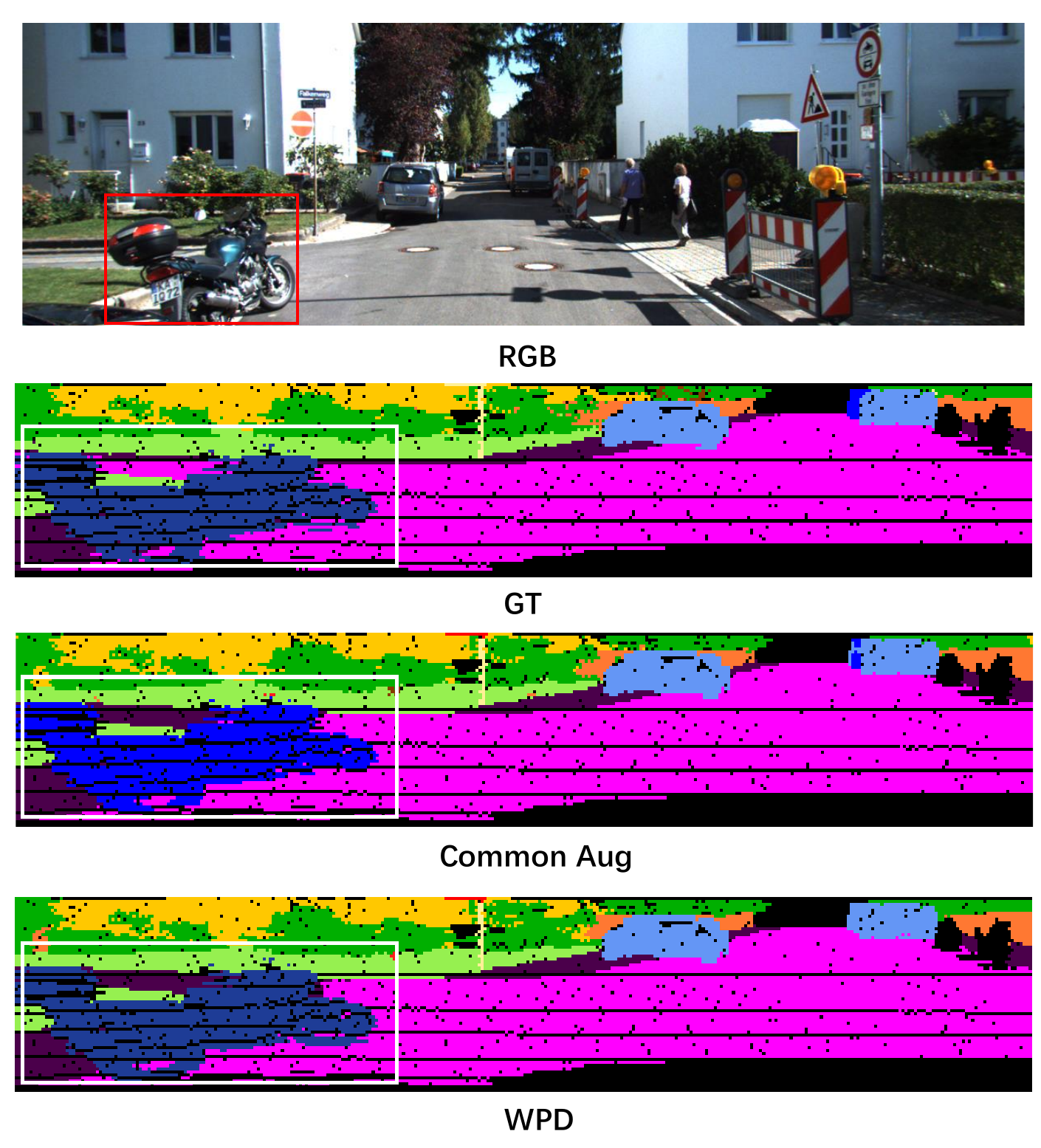}}
	\caption{Illustration of context reliance problem. With the common data augmentation, the predictions of MaskRange rely heavily on the context information. From the third row of the figure, the person and motorcycle on the road can be easily mis-classified into bicyclist and other-vehicle, respectively, since the bicyclist and other-vehicle tend to be on the road. With our WPD, MaskRange can correctly classify these objects and segment them precisely.} 
	\label{Context1}   
\end{figure}

\textbf{Context Reliance.} We compare the qualitative results of our MaskRange with common data augmentation and WPD. Figure \ref{Context1} shows the results on the validation set. 

\begin{figure}[htbp]
    \centering
	\includegraphics[width=1.0\textwidth]{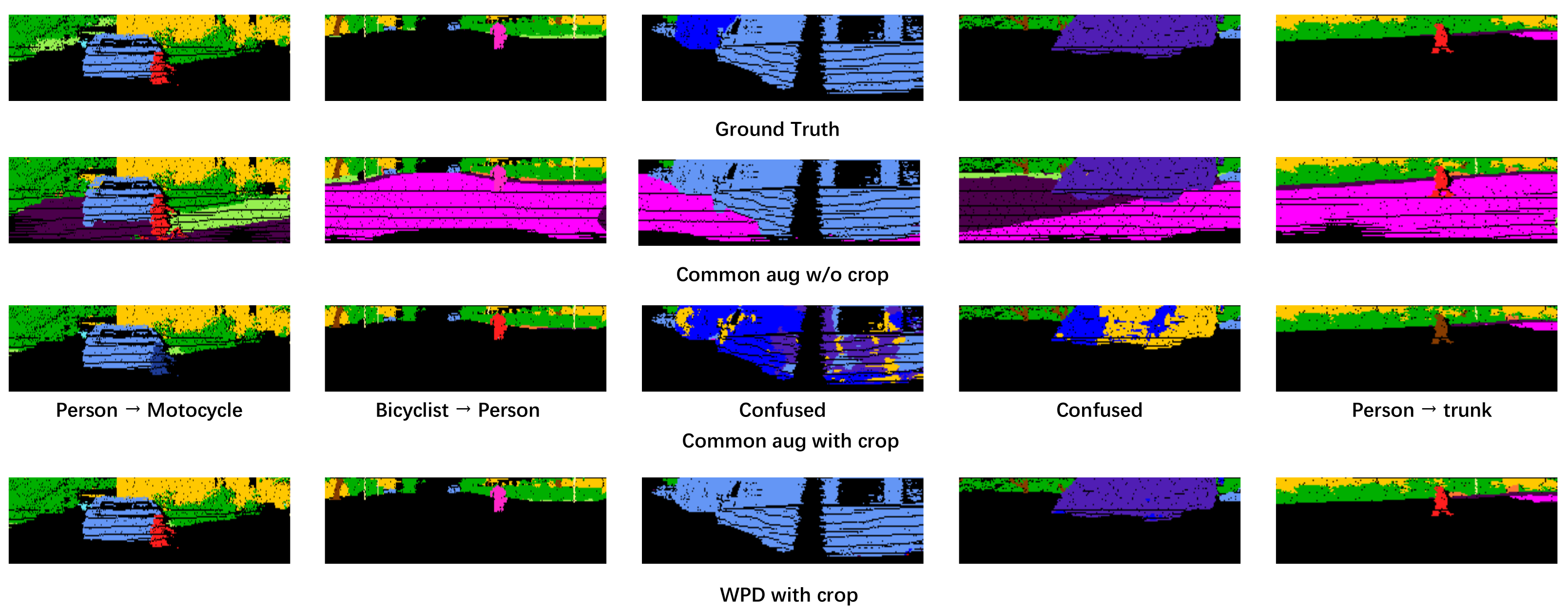} 
	\caption{The qualitative comparisons with ground removal.}
	\label{sup_4}   
\end{figure}

To further demonstrate the context reliance problem, we crop the original point cloud by dropping the points with $z (height) < -1.5$. In this way, most of ground points with high context information are dropped and we inference the rest points. Figure \ref{sup_4} shows that with common data augmentation scheme, MaskRange can correctly predict most of regions. However, with ground points removed, some regions are wrongly predicted, due to the loss of context information. Our WPD can effectively alleviate the context reliance problem and help MaskRange get reasonable predictions without ground points.

\section{Statistical Results}


We define the normalized weights for each class as: 
\begin{equation}
    w_{i}=
    \begin{cases}
        w_{s, i}, & \text{in semantic} \\
        w_{p, i}, & \text{in panoptic}
    \end{cases}
    ,\ \ w_{s, i}=\frac{\alpha_i}{\max\limits_{k }\alpha_k}, \ \ w_{p, i}=\frac{\alpha_i\beta_i}{\max\limits_{k }\alpha_k\beta_k},
\label{sem_pan_weights}
\end{equation}
where $\alpha_i = \frac{1}{f_i + \epsilon}$,  $\beta_i = \frac{ins_i}{sem_i}$, $k\in \{1, 2, ..., K\}$ and $K$ is the number of total classes. The $f_i$ is provided by SemanticKITTI \cite{behley2019semantickitti} and we obtain the class-wise statistical results by directly counting the number of $sem_i$ and $ins_i$ from training set. The detailed statistical results are shown in Table \ref{class-wise statistic}.
\vspace{10px}
\begin{table*}[htp]
\caption{Class-wise statistical results on SemanticKITTI \cite{behley2019semantickitti} training set. $f_i$ stands for the proportion of the class $i$. $ins_i$ and $sem_i$ are the number of instance and semantic segments of the class $i$, respectively. $\alpha_i$ is the semantic re-balance factor and $\beta_i$ is for panoptic. $w_{s, i}$ and $w_{p, i}$ are normalized weights of class $i$ for semantic and panoptic segmentation, respectively.}
\Huge
\resizebox{0.98\columnwidth}{!}{
\begin{tabular}
{c|l|ccccccccccccccccccc}
\toprule
Category & Method
& \begin{sideways} Car \end{sideways} 
& \begin{sideways} Bicycle \end{sideways} 
& \begin{sideways} Motorcycle \end{sideways} 
& \begin{sideways} Truck \end{sideways} 
& \begin{sideways} Other-vehicle \end{sideways} 
& \begin{sideways} Person \end{sideways} 
& \begin{sideways} Bicyclist \end{sideways} 
& \begin{sideways} Motorcyclist \end{sideways} 
& \begin{sideways} Road \end{sideways} 
& \begin{sideways} Parking \end{sideways} 
& \begin{sideways} Sidewalk \end{sideways} 
& \begin{sideways} Other-ground \end{sideways} 
& \begin{sideways} Building \end{sideways} 
& \begin{sideways} Fence \end{sideways} 
& \begin{sideways} Vegetation \end{sideways} 
& \begin{sideways} Trunk \end{sideways} 
& \begin{sideways} Terrain \end{sideways} 
& \begin{sideways} Pole \end{sideways} 
& \begin{sideways} Traffic-sign \end{sideways} \\
\midrule
\multirow{4}{*}{Semantic}
& $f_i$
& $4.26e^{-2}$ & $1.66e^{-4}$ & $3.98e^{-4}$ & $2.16e^{-3}$ & $1.81^e{-3}$ & $3.38e^{-4}$ & $1.27e^{-4}$ & $3.75e^{-5}$ & $1.99e^{-1}$ & $1.47e^{-2}$ & $1.44e^{-1}$ & $3.9e^{-3}$ & $1.33e^{-1}$ & $7.24e^{-2}$ & $2.67e^{-1}$ & $6.04e^{-3}$ & $7.81e^{-2}$ & $2.86e^{-3}$ & $6.16e^{-4}$ \\
& $\alpha_i$
& 22.93 & 857.56 & 715.11 & 315.96 & 356.25 & 747.62 & 887.22 & 963.89 & 5.01 & 63.62 & 6.9 & 203.88 & 7.48 & 13.63 & 3.73 & 142.15 & 12.64 & 259.37 & 618.97 \\
& $w_{s, i}$
& 0.02 & 0.89 & 0.74 & 0.33 & 0.37 & 0.78 & 0.92 & 1 & 0.01 & 0.07 & 0.01 & 0.21 & 0.01 & 0.01 & 0 & 0.15 & 0.01 & 0.27 & 0.64 \\
& $sem_i$
& 17784 & 3471 & 2872 & 2264 & 5063 & 4721 & 1176 & 552 & 19130 & 7705 & 18052 & 5257 & 17116 & 18751 & 19130 & 17124 & 18534 & 18617 & 13224 \\

\midrule
\multirow{4}{*}{Panoptic}
& $ins_i$
& 168431 & 6584 & 3444 & 2575 & 7499 & 8039 & 1492 & 559 & 19130 & 7705 & 18052 & 5257 & 17116 & 18751 & 19130 & 17124 & 18534 & 18617 & 13224 \\
& $\beta_i$
& 9.47 & 1.89 & 1.19 & 1.13 & 1.48 & 1.7 & 1.26 & 1.01 & 1 & 1 & 1 & 1 & 1 & 1 & 1 & 1 & 1 & 1 & 1 \\
& $\alpha_i\beta_i$
& 217.16 & 1620.79 & 850.98 & 357.04 & 527.24 & 1270.95 & 1117.9 & 973.53 & 5.01 & 63.62 & 6.9 & 203.88 & 7.48 & 13.63 & 3.73 & 142.15 & 12.64 & 259.37 & 618.97 \\
& $w_{p, i}$
& 0.13 & 1 & 0.53 & 0.22 & 0.33 & 0.78 & 0.69 & 0.6 & 0 & 0.04 & 0 & 0.13 & 0 & 0.01 & 0 & 0.09 & 0.01 & 0.16 & 0.38 \\
\bottomrule
\end{tabular}
}
\label{class-wise statistic}
\end{table*}
\vspace{10px}

For semantic segmentation, we define the classes with $w_{s, i} > t$ as long-tail classes. As for panoptic segmentation, the long-tail classes are those with $w_{p, i} > t$. Here, $t$ is the threshold to separate the long-tail and non-long-tail classes and is empirically set as $0.1$.

\begin{figure}[htbp]
    \centering
    \subfigure[]{\label{stat_1}
    \includegraphics[width=0.48\textwidth]{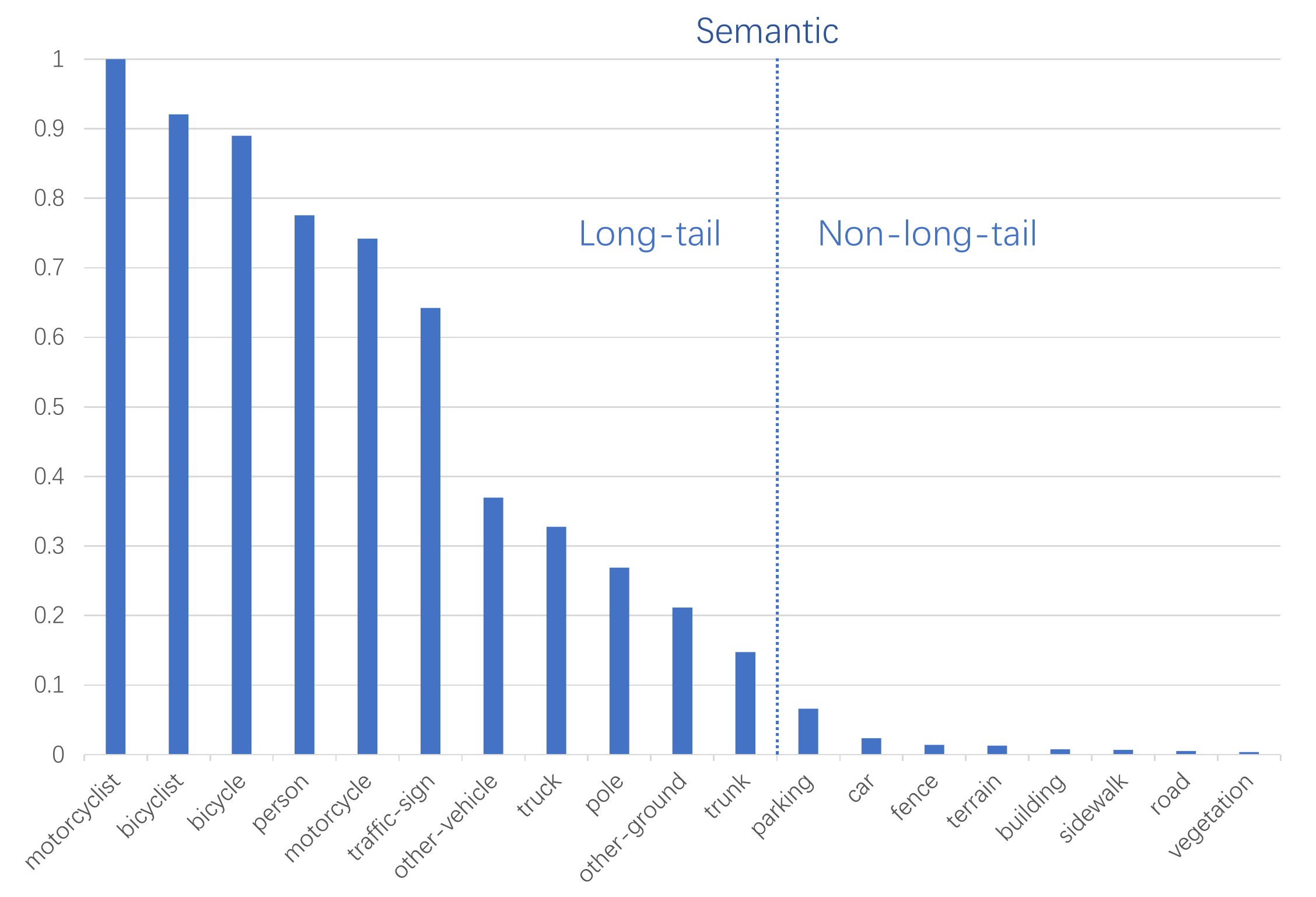}}
    \subfigure[]{\label{stat_2}
    \includegraphics[width=0.48\textwidth]{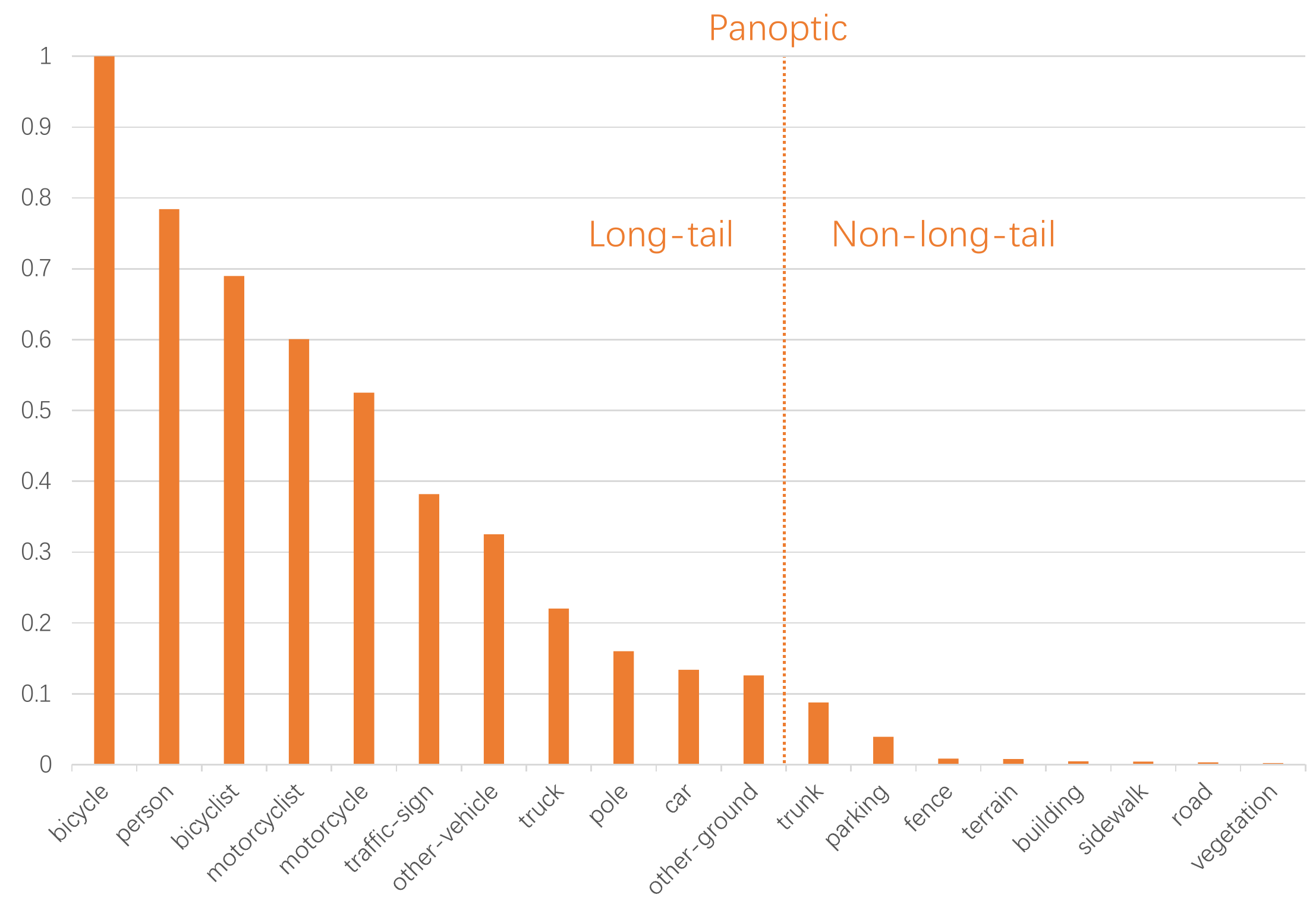}}
	\caption{Statistical results of class-wise normalized weights.} 
	\label{sup_statistic}   
\end{figure}

In this way, we set \textbf{motorcyclist}, \textbf{bicyclist}, \textbf{bicycle}, \textbf{person}, \textbf{motorcycle}, \textbf{traffic-sign}, \textbf{other-vehicle}, \textbf{truck}, \textbf{pole}, \textbf{other-ground}, \textbf{trunk} as long-tail classes for semantic sementation (Figure \ref{stat_1}) and \textbf{bicycle}, \textbf{person}, \textbf{bicyclist}, \textbf{motorcyclist}, \textbf{motorcycle}, \textbf{traffic-sign}, \textbf{other-vehicle}, \textbf{truck}, \textbf{pole}, \textbf{car}, \textbf{other-ground} as long-tail classes for panoptic segmentation (Figure \ref{stat_2}). 

\clearpage

\section{Data Augmentation details}
\subsection{WPD Probability}
The ``weighted" operation adds the probability to the ``paste" and ``drop". 

We only ``paste" long-tail classes with the probability $p_{i}$ for class $i$:
\begin{equation}
    p_{i}=
    \begin{cases}
        w_{s, i}-t, & \text{in semantic} \\
        w_{p, i}-t, & \text{in panoptic}
    \end{cases}.
\label{sem_probabilities}
\end{equation}

And we ``drop" the non-long-tail classes with the probability $d_{i}$:
\begin{equation}
    d_{i}=
    \begin{cases}
        t-w_{s, i}, & \text{in semantic} \\
        t-w_{p, i}, & \text{in panoptic}
    \end{cases}.
\label{pan_probabilities}
\end{equation}
Here, $t = 0.1$ is the threshold to separate the long-tail and non-long-tail classes.

\subsection{Ablation study}
We also do the ablation studies on the weighted paste, weighted drop and weighted paste drop. The results are presented in Table \ref{ablation_data}.

\begin{table}[htbp]
\caption{Ablation study for the data augmentation of LiDAR semantic segmentation on SemanticKITTI validation set. \textbf{WP}: weighted paste. \textbf{WD}: weighted drop. The \textbf{WPD}: weighted paste with drop. }
\begin{center}
\begin{tabular}{c|c c c|c}
\hline
 & WP & WD & WPD & mIoU (\%)\\
\hline\hline
\multirow{3}*{MaskRange} 
 & \checkmark &  &  & 66.09\\
 &  & \checkmark &  & 64.80\\
 &  &  & \checkmark & \textbf{67.77}\\
\hline
\end{tabular}
\end{center}
\label{ablation_data}
\end{table}

\subsection{Visual comparisons}

\begin{figure}[htbp]
    \centering
	\includegraphics[width=1.0\textwidth]{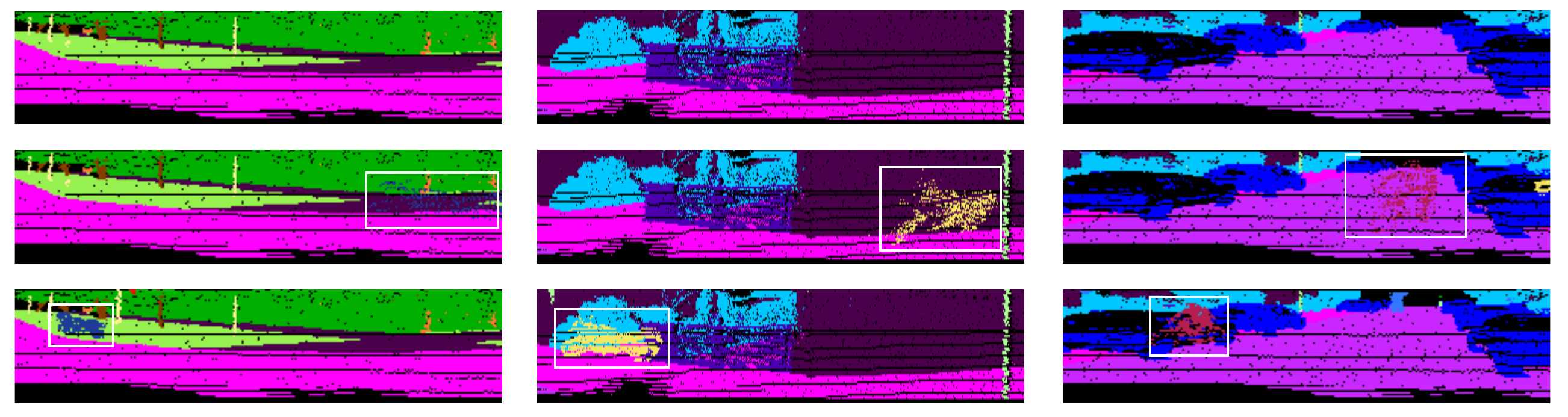} 
	\caption{Sparsity problem visualization. First to third rows are original image, instance paste \cite{xu2021rpvnet} and our paste, respectively. The regions in white boxes are augmented areas. Instance paste add the objects above the random ground points, which may put the far away objects to the near places and thus result in the sparsity problem. Our method pastes the actual scanning points, which inherently avoids this problem.} 
	\label{sup_sparsity}   
\end{figure}

\begin{figure}[htbp]
    \centering
	\includegraphics[width=1.0\textwidth]{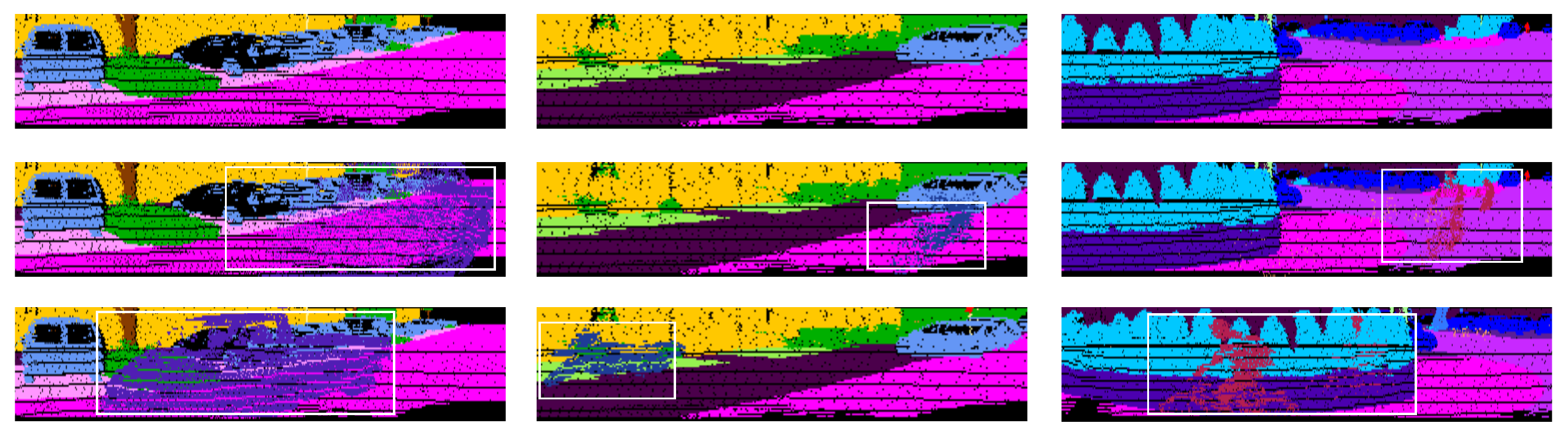}
	\caption{Orientation problem visualization. First to third rows are original image, instance paste \cite{xu2021rpvnet} and our paste, respectively. The regions in white boxes are augmented areas. Instance paste does not consider the orientation problem by putting objects with random orientations, which may make model confused. Our method pastes the actual scanning points, which inherently avoids this problem.} 
	\label{sup_orientation}   
\end{figure}

\begin{figure}[htbp]
    \centering
    \subfigure[raw image, Mix3D, our WPD]{\label{supwpd_3}
    \includegraphics[width=0.96\textwidth]{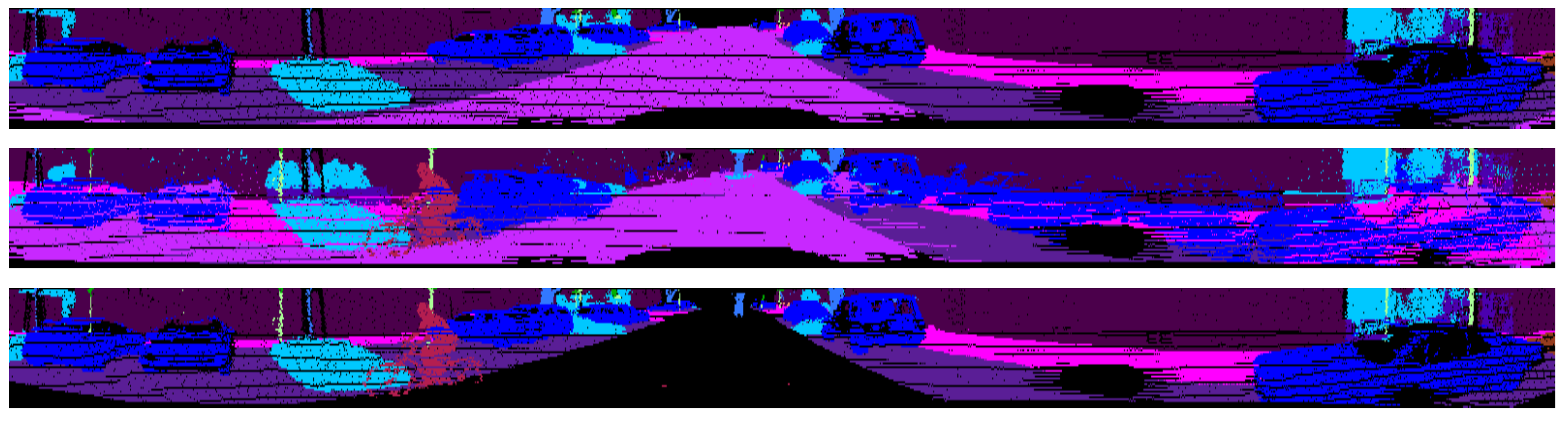}}
    \subfigure[raw image, Mix3D, our WPD]{\label{supwpd_4}
    \includegraphics[width=0.96\textwidth]{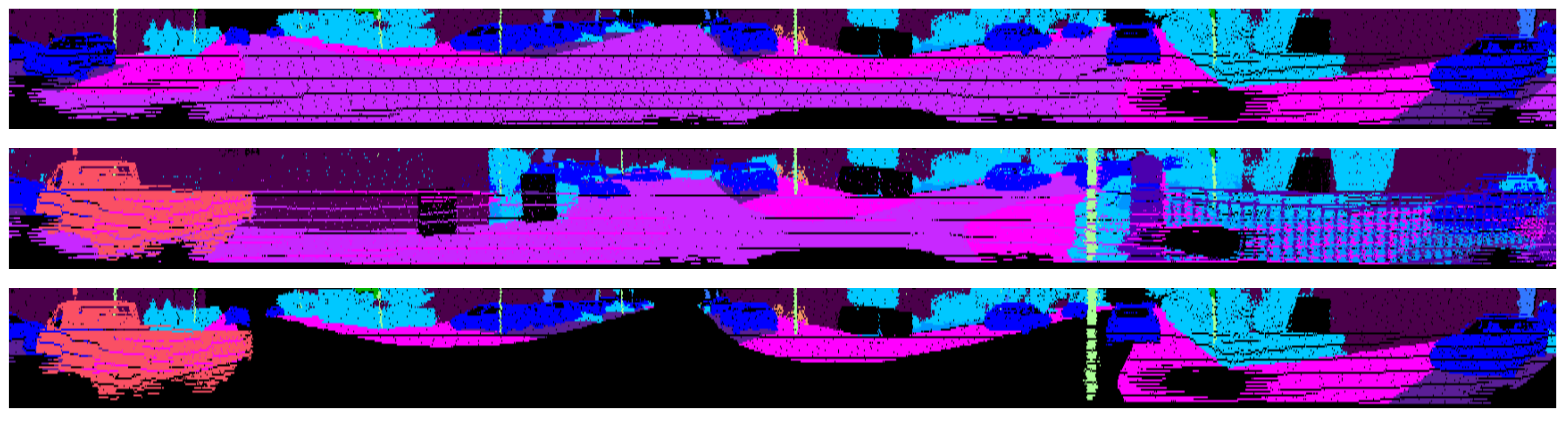}}
    \subfigure[raw image, Mix3D, our WPD]{\label{supwpd_5}
    \includegraphics[width=0.96\textwidth]{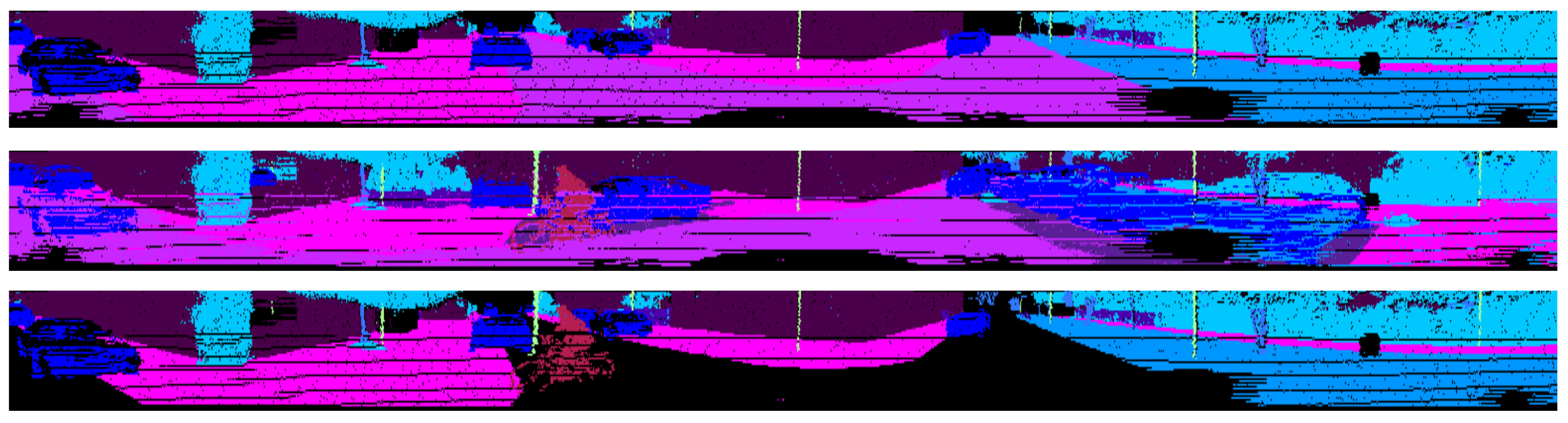}}
	\caption{Mix3d \cite{nekrasov2021mix3d} vs. WPD.} 
	\label{sup_wpd}   
\end{figure}

We sample some visual results to show the sparsity and orientation problems (Figure \ref{sup_sparsity} and \ref{sup_orientation}). We also compare the augmented range images with Mix3D \cite{nekrasov2021mix3d} (Figure \ref{sup_wpd}).

\clearpage

\section{Class Balance vs. Unified Balance}
\begin{table}[htbp]
\caption{Ablation study of the re-balance strategies on the SemanticKITTI \cite{behley2019semantickitti} validation set.}
\label{ablation_balance}
\begin{center}
\begin{tabular}{c|c c c|c}
\hline
 & N & C & U & mIoU (\%)\\
\hline\hline
\multirow{3}*{Semantic} & \checkmark &  &  & 64.43\\
 &  & \checkmark &  & \textbf{66.74}\\
 &  &  & \checkmark & \textbf{66.74}\\
\hline
 & N & C & U & PQ (\%)\\
\hline
\multirow{3}*{Panoptic} & \checkmark &  &  & 48.33\\
 &  & \checkmark &  & 49.43\\
 &  &  & \checkmark & \textbf{51.15}\\
\hline
\end{tabular}
\end{center}
\end{table}

We investigate the effectiveness of the re-balance designs by comparing three different strategies:

\textbf{No balance (N)} : Paste + Drop + focal loss 

\textbf{Class balance (C)} : Weighted + Paste + Drop + focal loss + $\alpha_i$

\textbf{Unified balance (U)} : Weighted + Paste + Drop + focal loss + $\alpha_i$ + $\beta_i$

We report the results on both semantic and panoptic segmentation in Table \ref{ablation_balance}. Compared with no-balance strategy, the class-balance strategy shows better performance on both two tasks. However, the class-balance strategy is only designed for semantic segmentation. This motivates us to deal with the imbalance problem of semantic and panoptic segmentation in a unified manner considering that our MaskRange is a unified architecture. Thus, we propose the unified-balance strategy, which achieves better performance on panoptic segmentation and the same performance as the class-balance strategy on semantic segmentation.

\section{Post-processing}
The projection-based methods suffer from the ``shadow" effect \cite{milioto2019rangenet++}, which is caused by the blurry CNN mask and the many-to-one mapping. To alleviate this ``shadow" effect, a KNN (K-nearest-neighbor) post-processing module \cite{milioto2019rangenet++} is proposed and we adopt this post-processing as well. Furthermore, a simple but effective post-processing method for our MaskRange can be inserted to the previous stage of the KNN post-processing. 

\begin{figure}[htb]
\centering
\subfigure[Before temporal filter.]{
\includegraphics[width=0.48\textwidth]{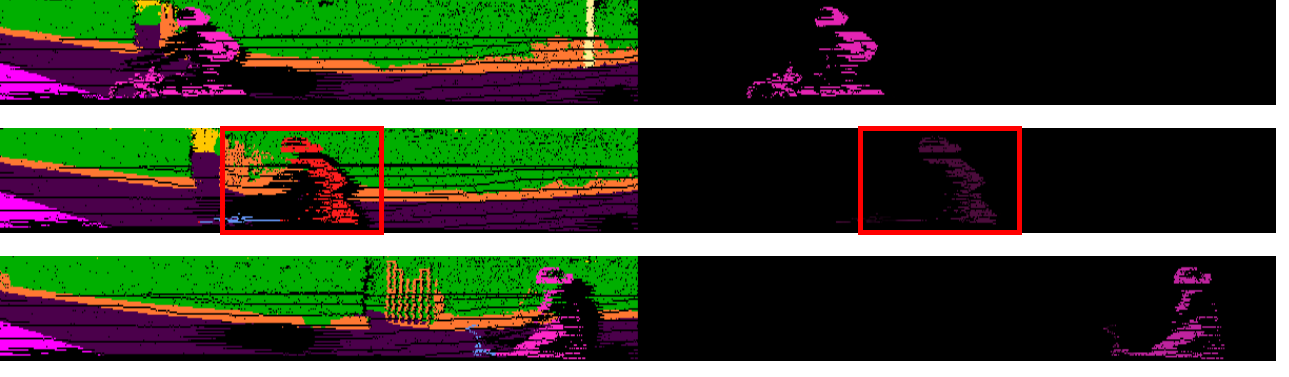}
\label{tp_a}
}
\subfigure[After temporal filter.]{
\includegraphics[width=0.48\textwidth]{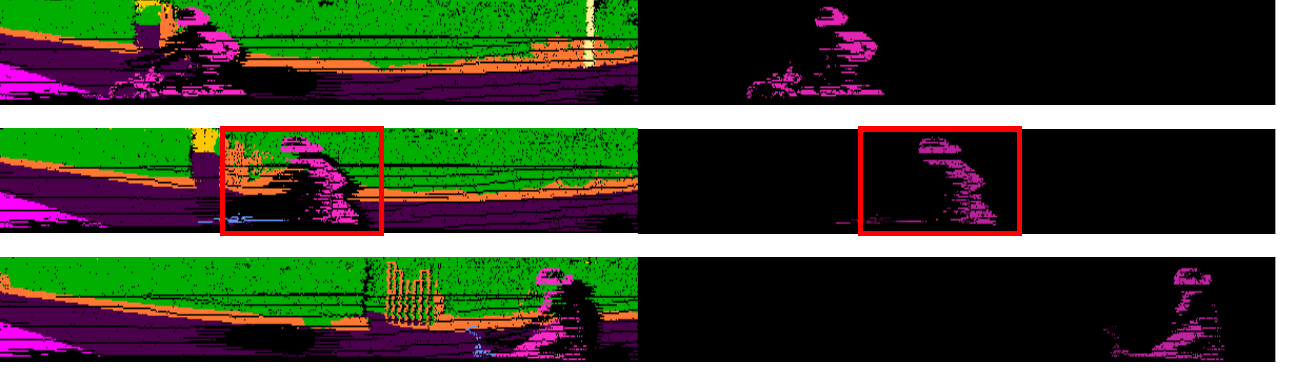}
\label{tp_b}
}
\caption{The illustration of the temporal filter post-processing. The semantic predictions and mask predictions (from the same query) from three continuous frames are shown in figure \ref{tp_a} and \ref{tp_b}. As shown in figure \ref{tp_a}, a bicyclist is mis-classified to person because of the low confidence of its mask prediction. After the temporal filter post-processing (figure \ref{tp_b}), the confidence is improved by utilizing the temporal consistency and the corresponding region is corrected.}
\label{post}
\end{figure}

In real-world applications, LiDAR frames are usually segmented frame by frame. As shown in Figure \ref{post}, some predictions can be mis-classified and result in a ``flickering" effect over time. The ``flickering" is caused by the largely different observation views and also exists in per-pixel classification methods. 

Different from per-pixel classification methods, our MaskRange can aggregate the prediction errors due to the introduced mask. If a query is mis-classified, the whole region will be wrongly predicted. At the inference stage, the object queries are fixed and the classification results of these queries tend to be consistent in continuous frames. We can utilize this property to further improve the performance. 

We notate the classification output of query $i$ in the current frame $t$ as $\hat{C}_{t, i}$. We set a temporal window centered at frame $t$ with the size $K + L + 1$, where $K$ is the number of previous frames and $L$ is the number of future frames. To alleviate the ``flickering" effect, we simply apply a temporal filter to the class predictions of these queries within the window. For example, the average filter can be computed as:
\begin{equation}
    \bar{\hat{C}}_{t, i} = \frac{1}{L+K+1}\sum_{t}\hat{C}_{t, i}.
\end{equation}

Note that our filtering post-processing is easy to implement and takes almost negligible runtime and memory cost. It utilizes the continuous inputs during inference when applied to real word scene understanding, free of sequential training data. Our filtering post-processing can be applied together with KNN post-processing and does not rely on future information (simply set $L$ as $0$). 

\section{More Results}
Table \ref{post_appendix} shows the results of our temporal filter post-processing.

\begin{table}[htp]
\caption{Results of MaskRange with temporal filter post-processing on the SemanticKITTI \cite{behley2019semantickitti} test set.}
\centering
 \begin{tabular}{|c|c|} 
 \hline
 Methods & Semantic (mIoU) \\ 
 \hline
 KNN & 66.10  \\
 \hline
 KNN+temporal & 66.34 \\
 
 \hline
 \end{tabular}
\label{post_appendix}
\end{table}

The class-wise panoptic results are presented in Table \ref{tab:panoptic_validationset}.

\begin{table*}[htp]
\setlength\tabcolsep{3.7pt}
\begin{center}
\caption{Class-wise PQ scores of LiDAR panoptic segmentation on the SemanticKITTI \cite{behley2019semantickitti} test set. R.Net, P.P, KPC refer to RangeNet++, Point Pillars, KPConv respectively.}
        \label{tab:panoptic_validationset}
        \footnotesize
        \resizebox{0.98\columnwidth}{!}{
        \begin{tabular}{l|ccccccccccccccccccc|c}
            \toprule
            Method & \begin{sideways}car\end{sideways} & \begin{sideways}truck\end{sideways} & \begin{sideways}bicycle\end{sideways} & \begin{sideways}motorcycle\end{sideways} & \begin{sideways}other vehicle\end{sideways} & \begin{sideways}person\end{sideways} & \begin{sideways}bicyclist\end{sideways} & \begin{sideways}motorcyclist\end{sideways} & \begin{sideways}road\end{sideways} & \begin{sideways}sidewalk\end{sideways} & \begin{sideways}parking\end{sideways} & \begin{sideways}other ground\end{sideways} & \begin{sideways}building\end{sideways} & \begin{sideways}vegetation\end{sideways} & \begin{sideways}trunk\end{sideways} & \begin{sideways}terrain\end{sideways} & \begin{sideways}fence\end{sideways} & \begin{sideways}pole\end{sideways} & \begin{sideways}traffic sign\end{sideways} & PQ \\
            \midrule
            R.Net~\cite{milioto2019rangenet++}+ P.P.~\cite{lang2019pointpillars} & 66.9 & 6.7 & 3.1 & 16.2 & 8.8 & 14.6 & 31.8 & 13.5 & 90.6 & 63.2 & 41.3 & 6.7 & 79.2 & 71.2 & 34.6 & 37.4 & 38.2 & 32.8 & 47.4 & 37.1\\
            KPC~\cite{thomas2019kpconv} + P.P.~\cite{lang2019pointpillars} & 72.5 & 17.2 & 9.2 & 30.8 & 19.6 & 29.9 & 59.4 & 22.8 & 84.6 & 60.1 & 34.1 & 8.8 & 80.7 & 77.6 & 53.9 & 42.2 & 49.0 & 46.2 & 46.8 & 44.5\\
            LPSAD~\cite{9340837} & 76.5 & 7.1 & 6.1 & 23.9 & 14.8 & 29.4 & 29.7 & 17.2 & 90.4 & 60.1 & 34.6 & 5.8 & 76.0 & 69.5 & 30.3 & 36.8 & 37.3 & 31.3 & 45.8 & 38.0 \\
            PanopticTrackNet~\cite{hurtado2020mopt} &70.8 & 14.4 & 17.8 & 20.9 & 27.4 &34.2 & 35.4& 7.9 & 91.2 & 66.1 & 50.3 & 10.5 & 81.8 & 75.9 & 42.0 & 44.3 & 42.9 & 33.4 & 51.1 & 43.1 \\
            Panoster~\cite{gasperini2020panoster} &  84.0 & 18.5 & 36.4 & 44.7 & 30.1 & 61.1 & \textbf{69.2} & 51.1 & 90.2 & 62.5 & 34.5 & 6.1 & 82.0 & 77.7 & \textbf{55.7} & 41.2 & 48.0 & 48.9 & 59.8 & 52.7\\
            EfficientLPS~\cite{sirohi2021efficientlps} & \textbf{85.7} & 30.3 & \textbf{37.2} & \textbf{47.7} & \textbf{43.2} & \textbf{70.1} & 66.0 & 44.7 & \textbf{91.1} & 71.1 & \textbf{55.3} & \textbf{16.3} & \textbf{87.9} & \textbf{80.6} & 52.4 & \textbf{47.1} & \textbf{53.0} & 48.8 & 61.6 & \textbf{57.4} \\
            \midrule
            MaskRange (ours) & $70.8$ & \textbf{37.8} & $15.8$ & $38.4$ & $37.2$ & $42.5$ & $59.5$ & \textbf{57.3} & $88.7$ & \textbf{71.3} & $50.1$ & $9.76$ & $87.4$ & $79.1$ & $54.3$ & $44.1$ & $47.8$ & \textbf{52.3} & \textbf{65.2} & $53.1$ \\
            \bottomrule
        \end{tabular}
        }
    \end{center}
\end{table*}
\end{document}